\documentclass[conference]{IEEEtran}
\IEEEoverridecommandlockouts
\usepackage{cite}
\usepackage{amsmath,amssymb,amsfonts}
\usepackage{algorithmic}
\usepackage{graphicx}
\usepackage{textcomp}
\usepackage{xcolor}
\def\BibTeX{{\rm B\kern-.05em{\sc i\kern-.025em b}\kern-.08em
    T\kern-.1667em\lower.7ex\hbox{E}\kern-.125emX}}
    
\usepackage{adjustbox}
\usepackage{caption}
\usepackage{microtype}
\usepackage{booktabs}
\usepackage{multirow} 
\usepackage{marvosym}
\usepackage{cite}
\usepackage{multirow}

\usepackage{amsthm}

\newcommand{\refdef}[1]{Definition~\ref{#1}}

\newcommand{\refcorollary}[1]{Corollary~\ref{#1}}
\newcommand{\reftheorem}[1]{Theorem~\ref{#1}}
\newcommand{\refeq}[1]{Eq.~\eqref{#1}}

\usepackage{listings}
\usepackage{ifxetex}
\usepackage{fancyhdr}
\usepackage{hyperref}
\usepackage{graphicx}
\usepackage{wrapfig}
\usepackage{amsmath}
\usepackage{amsfonts}
\usepackage{amssymb}
\usepackage{listings}
\usepackage{setspace}
\usepackage{wrapfig}
\usepackage{tablefootnote}
\usepackage{multirow}
\usepackage{booktabs}
\usepackage[section]{placeins}
\usepackage{color}
\usepackage{amsfonts}
\usepackage{ulem}
\usepackage{float}
\usepackage{dsfont}
\usepackage{cleveref}

\newcommand{\mat}[1]{\mathbf{#1}}
\newcommand{\set}[1]{\mathcal{#1}}
\newcommand{\pnorm}[1]{\lVert{#1}\rVert}

\DeclareMathOperator*{\loss}{{\ell}}
\DeclareMathOperator*{\dist}{{d}}

\DeclareMathOperator*{\trace}{{trace}}

\DeclareMathOperator*{\Var}{Var}

\DeclareMathOperator*{\diag}{diag}
\DeclareMathOperator*{\sign}{sign}

\newcommand{\N}{\ensuremath{\mathcal{N}}} % normal distribution
\newcommand{\U}{\ensuremath{\mathcal{U}}} % uniform distribution
\newcommand{\I}{\mat{\mathbb{I}}}
\newcommand{\E}{\mathbb{E}}
\newcommand{\RN}{\mathbb{R}}

\newcommand{\Lagrange}{\mathcal{L}}
\DeclareMathOperator*{\densitythreshold}{\ensuremath{\delta}}
\newcommand{\density}{\ensuremath{p}}
\newcommand{\densityestimator}{\ensuremath{\hat{p}}}

\DeclareMathOperator*{\regularization}{\ensuremath{{\theta}}}
\newcommand{\x}{\ensuremath{\vec{x}}}

\newcommand{\xorig}{\ensuremath{\vec{x}_{\text{orig}}}}
\newcommand{\yorig}{\ensuremath{y_{\text{orig}}}}
\newcommand{\setx}{\ensuremath{\set{X}}}
\newcommand{\sety}{\ensuremath{\set{Y}}}

\newcommand{\xcf}{\ensuremath{\vec{x}_\text{cf}}}
\newcommand{\xcfNew}{\ensuremath{\vec{x}_\text{cf}'}}
\newcommand{\CF}[2]{\ensuremath{\text{CF}(#1,#2)}}
\newcommand{\ycf}{\ensuremath{y'}}

\newcommand{\w}{\ensuremath{\vec{w}}}

\newcommand{\dimsym}{d}
\newcommand{\classifier}{\ensuremath{h}}

\newcommand{\mask}{\ensuremath{\vec{m}}}
\newcommand{\prob}{\ensuremath{\mathbb{P}}}
\newcommand{\CovMat}{\ensuremath{\mat{\Sigma}}}
\newcommand{\PerturbationDensity}{\ensuremath{p_{\epsilon}}}

\newtheorem{theorem}{Theorem}
\newtheorem{definition}{Definition}

\newtheorem{corollary}[theorem]{Corollary}
\newtheorem{remark}{Remark}

\usepackage{todonotes}
\begin{document}

\title{Evaluating Robustness of\\ Counterfactual Explanations
\thanks{We gratefully acknowledge funding from the VW-Foundation for the project \textit{IMPACT} funded in the frame of the funding line \textit{AI and its Implications for Future Society},
fundings from the federal state government
of North Rhine-Westphalia (NRW) for the projects \textit{Bias von KI-Modelle bei der Informationsbildung und deren Implikationen in der Wirtschaft} and the research training group \textit{Dataninja} (Trustworthy AI for Seamless Problem Solving: Next Generation Intelligence Joins Robust Data Analysis), and
funding from the German Federal Ministry of Education and Research (BMBF) through the project \textit{TiM} (05M20PBA).
}
}

\author{\IEEEauthorblockN{1\textsuperscript{st} Andr\'e Artelt}
\IEEEauthorblockA{\textit{CITEC} \\
\textit{Bielefeld University}\\
Bielefeld, Germany \\
aartelt@techfak.de}
\and
\IEEEauthorblockN{2\textsuperscript{nd} Valerie Vaquet}
\IEEEauthorblockA{\textit{CITEC} \\
\textit{Bielefeld University}\\
Bielefeld, Germany \\
vvaquet@techfak.de}
\and
\IEEEauthorblockN{3\textsuperscript{rd} Riza Velioglu}
\IEEEauthorblockA{\textit{CITEC} \\
\textit{Bielefeld University}\\
Bielefeld, Germany \\
rvelioglu@techfak.de}
\and
\IEEEauthorblockN{4\textsuperscript{th} Fabian Hinder}
\IEEEauthorblockA{\textit{CITEC} \\
\textit{Bielefeld University}\\
Bielefeld, Germany \\
fhinder@techfak.de}
\and
\IEEEauthorblockN{5\textsuperscript{th} Johannes Brinkrolf}
\IEEEauthorblockA{\textit{CITEC} \\
\textit{Bielefeld University}\\
Bielefeld, Germany \\
jbrinkro@techfak.de}
\and
\IEEEauthorblockN{6\textsuperscript{th} Malte Schilling}
\IEEEauthorblockA{\textit{CITEC} \\
\textit{Bielefeld University}\\
Bielefeld, Germany \\
mschilli@techfak.de}
\and
\IEEEauthorblockN{7\textsuperscript{th} Barbara Hammer}
\IEEEauthorblockA{\textit{CITEC} \\
\textit{Bielefeld University}\\
Bielefeld, Germany \\
bhammer@techfak.de}
}

\maketitle

\begin{abstract}
Transparency is a fundamental requirement for decision making systems when these should be deployed in the real world. It is usually achieved by providing explanations of the system's behavior. A prominent and intuitive type of explanations are counterfactual explanations. Counterfactual explanations explain a behavior to the user by proposing actions---as changes to the input---that would cause a different (specified) behavior of the system.
However, such explanation methods can be unstable with respect to small changes to the input---i.e. even a small change in the input can lead to huge or arbitrary changes in the output and of the explanation.
This could be problematic for counterfactual explanations,  as two similar individuals might get very different explanations. Even worse, if the recommended actions differ considerably in their complexity, one would consider such unstable (counterfactual) explanations as individually unfair.

In this work, we formally and empirically study the robustness of counterfactual explanations in general, as well as under different models and different kinds of perturbations. Furthermore, we propose that plausible counterfactual explanations can be used instead of closest counterfactual explanations to improve the robustness and consequently the individual fairness of counterfactual explanations.
\end{abstract}


\section{Introduction}
Transparency is a fundamental building block for ethical artificial intelligence (AI) and machine learning (ML) based decision making systems. In particular, the increasing use of automated decision making systems in the real world~\cite{predictivepolicing,creditscoresunfair} has strengthened the demand for trustworthy systems.
The criticality of transparency was also recognized recently by policy makers which resulted in legal regulations like the EU's GDPR~\cite{gdpr} that grants the user a right to an explanation. 
Therefore, explainability and transparency of AI and ML methods
has become an important research focus in the last years~\cite{explainingexplanations, explainingblackboxmodelssurvey, surveyxai, explainableartificialintelligence}. Nowadays, there are several technologies available for explaining  ML models~\cite{explainingblackboxmodelssurvey,molnar2019}.
One specific family of methods are model-agnostic methods~\cite{explainingblackboxmodelssurvey,modelagnosticinterpretability}, which are not tailored to a particular model or representation and thus  applicable to a wide range of ML models. 
%In particular, ``truly'' model-agnostic methods do not need access to the training data or model internals. It is sufficient to have an interface for passing data points to the model and obtaining the output of the model - the underlying model it self is viewed as a black-box.

Examples of model-agnostic methods are feature interaction methods~\cite{featureinteraction}, feature importance methods~\cite{featureimportance}, partial dependency plots~\cite{partialdependenceplots}, and  methods that approximate the model locally by an explainable model~\cite{lime2016,decisiontreecounterfactual}. These methods explain the models by using {\it input features}\ as the vocabulary.

In contrast, in {\it example-based explanations}\ ~\cite{casebasedreasoning} a prediction or behavior is explained by a (set of) data points. % are a different type of model-agnostic explanation.
Instances of example-based explanations are prototypes \& criticisms~\cite{prototypescriticism}, influential instances~\cite{influentialinstances}, and counterfactual explanations~\cite{counterfactualwachter}. A counterfactual explanation states a change to the original input that results in a different (specific) prediction or behavior of the decision making system--- \textit{what needs to be different in order for the system's prediction to change?} Such an explanation is considered to be intuitive and useful because it proposes changes to achieve a desired outcome, i.e.\ it provides {\it actionable feedback}~\cite{molnar2019,counterfactualwachter}. Furthermore, there exist strong evidence that explanations by humans are often counterfactual in nature~\cite{ijcai2019-876} and preferred. Our focus will be on counterfactual explanations.

However, it has also been shown that many explanation methods are vulnerable to adversarial attacks~\cite{FairwashingExplanations,FoolingNeuralNetworkInterpretations,DBLP:journals/corr/abs-1806-08049}---i.e. an explanation (such as a saliency map or feature importances) can be (arbitrarily) changed by applying only small perturbations to the original sample that should be explained. This instability of explanations applies to counterfactual explanations as well~\cite{IssuesPostHocCounterfactualExplanations}. As the necessity of local stability of explanations is widely accepted~\cite{LocalitySurogatesPostHocInterpretability,IssuesPostHocCounterfactualExplanations,FairwashingExplanations,FoolingNeuralNetworkInterpretations,RobustInterpretabilityNeuralNetworks}, we face the challenge how to compute stable and robust counterfactual explanations which is still an open-research problem~\cite{verma2020counterfactual}.

Missing stability and robustness of explanations can lead to unfair explanations and thus compromise the system's trustworthiness~\cite{RobustInterpretabilityNeuralNetworks,FairwashingExplanations}. In particular, this holds true for counterfactual explanations because their explanations lead to actions: since these can be interpreted as proposed actions in order to achieve a desired goal, it is problematic, if the proposed actions differ considerably between individuals without a clear reason.
\textit{For instance consider the scenario of loan application where two (financially) similar individuals applied for a loan and both applications got rejected. However, the counterfactual explanation both people get are highly different---one applicant only needs a minor increase in monthly income, while the other applicant is told that a major increase in monthly income as well as some other aspects are required for getting the loan application accepted.}
Thus, robustness of counterfactual explanations is closely related to individual fairness of counterfactual explanations.

%\todo{Remove this block? However, I think some words on fairness are necessary since we apply the concept of individual fairness to counterfactual explanations}
%\todo{Agreed: This sticks out a little bit - might boil it down, shorten it. What is the goal of this paragraph? Introducing related work to fairness? Maybe start with that: Fairness of ML based decision making systems has been a topic of research in itself, mostly driven by prominent failures ... As a consequence, several ...}
Note that fairness of ML based decision making systems itself has been already studied for some time. Especially because of prominent failures of ML system in critical situations like predictive policing~\cite{predictivepolicing} and loan approval~\cite{creditscoresunfair}. Many of those failures, where the systems exhibit unethical behavior, deal with predictions that are correlated to sensitive attributes such as race, gender, etc.
As a consequence, several (formal) fairness criteria, concerning the predictions and behaviors of ML systems, have been proposed~\cite{BiasFairnessMLSurvey,FairnessMLSurvey}.
A large number of these criteria deal with the dependency between a sensitive attribute/feature and the response variable (prediction/behavior of the system)---these are known as group-based fairness criteria.
Prominent examples are~\cite{FairnessMLSurvey} demographic parity, equalized odds, predictive rate parity, or causal independence. However, there are fundamental limitations for an efficient 
implementation of fair models, since some intuitively relevant fairness criteria can be contradictory in specific examples.
Another formalization of fairness focuses on individual (rather than group) fairness~\cite{FairnessThroughAwareness}. The idea behind individual fairness is to treat similar individuals similarly---this notion resembles the concept of individual fairness from other scientific disciplines~\cite{FairnessThroughAwareness,IndividualVsGroupFairness}. Despite its appealing intuition, a major challenge of individual fairness constitutes in its formalization---i.e.\ given two individuals, how  can we score their mutual similarity? 

Despite its relevance, robustness of explanations has only recently started being investigated in first approaches~\cite{10.1145/3351095.3372836,DBLP:journals/corr/abs-1806-08049}, although it is well known that robustness and fairness concerning the predictions and behavior of ML models are closely related to each other~\cite{BiasRobustnessFairML,FairnessThroughRobustness}.
Consequently, only a few papers study the fairness of explanations itself.
For instance the authors of~\cite{DBLP:journals/corr/abs-1909-03166} propose a group fairness criteria ``Equalizing Recourse'' which aims for ensuring that the same opportunities hold for feasible recourse (as suggested by an explanation) across different groups.
The authors of~\cite{vonkuegelgen2021fairness} propose and study group fairness and individual fairness of causal recourse - i.e. recourse/explanations that are obtained from a given structural cause model (SCM).

\paragraph*{Our Contributions}
In this work, we propose a formalization of robustness of counterfactual explanations and investigate its formal mathematical properties and its experimental behavior for a few  popular models. We propose to use plausible counterfactuals instead of closest counterfactual explanations because we find evidence that the plausible counterfactual are more robust. Furthermore, we also relate our robustness definition to individual fairness and argue that our robustness definition can be interpreted as a definition of individual fairness and thus all our findings on robustness transfer to individual fairness of counterfactual explanations as well.
%Nice!

% change to Section-\ref{} (capital S), see http://www.ctan.org/tex-archive/macros/latex/contrib/IEEEtran/IEEEtran_HOWTO.pdf
To address these goals, first,  we briefly review counterfactual explanations in Section~\ref{sec:foundations}. Next, we formally define our notion of robustness of counterfactual explanations in Section~\ref{sec:counterfactual:fairness}. In Section~\ref{sec:counterfactuals:robustness}, we  study formal properties of robustness of counterfactual explanations (Section~\ref{sec:counterfactuals:robustness}), we propose the concept to use plausible instead of closest counterfactual explanations for improving the robustness of the explanations and also show how our robustness definition relates to individual fairness of counterfactual explanations (Section~\ref{sec:individualfairness}). We empirically evaluate the robustness of closest and plausible counterfactual explanations in Section~\ref{sec:experiments}. Finally, we give a summary and outlook in Section~\ref{sec:conclusion}.

\section{Foundations}\label{sec:foundations}
\subsection{Counterfactual Explanations}\label{sec:foundation:counterfactuals}
Counterfactual explanations (often just called counterfactuals) contrast samples by counterparts with minimum change of the appearance but different class label~\cite{counterfactualwachter,molnar2019} and can be formalized as follows: 
\begin{definition}[(Closest) Counterfactual explanation~\cite{counterfactualwachter}]\label{def:counterfactual}
Assume a prediction function $\classifier:\RN^\dimsym \to \sety$ is given. Computing a counterfactual $\xcf \in \RN^\dimsym$ for a given input $\xorig \in \RN^\dimsym$ is phrased as an optimization problem:
\begin{equation}\label{eq:counterfactualoptproblem}
\underset{\xcf \,\in\, \RN^\dimsym}{\arg\min}\; \loss\big(\classifier(\xcf), \ycf\big) + C \cdot \regularization(\xcf, \xorig)
\end{equation}
where $\loss(\cdot)$ denotes a suitable loss function, $\ycf$ the requested prediction, and  $\regularization(\cdot)$  a penalty term for deviations of $\xcf$ from the original input $\xorig$. $C>0$ denotes the regularization strength.
We refer to a counterfactual given $\xorig$ and desired class $\ycf$ as any solution of the optimization problem.
\end{definition}
In this work we assume that data comes from a real-vector space and continuous optimization is possible. In this context~\cite{counterfactualcomputationsurvey}, two common regularizations are the weighted Manhattan distance and the generalized $\l_2$ distance.
Depending on the model and the choice of $\loss(\cdot)$ and $\regularization(\cdot)$, the final optimization problem might be differentiable or not. If it is differentiable, we can use a gradient-based optimization algorithm like BFGS. Otherwise, we can rely on black-box optimization algorithms for continuous optimization like Nelder-Mead.
Note that counterfactuals need not be unique due to the possible existence of local optima of the optimization problem. We will deal with popular but simple models in the following where this is not the case. In general, we assume that an ordering of counterfactuals is induced by the respective optimization algorithm. 
Given input $\xorig$ and desired label $\ycf$, the counterfactual provided by the algorithm at hand is referred to as $\xcf=\CF{\xorig}{\ycf}$.
 
While the formalization of the optimization problem~\refeq{eq:counterfactualoptproblem} is model agnostic (i.e. it does not make any assumptions on the model $\classifier(\cdot)$), it can be beneficial to rewrite the optimization problem~\refeq{eq:counterfactualoptproblem} in constrained form~\cite{counterfactualcomputationsurvey}:
\begin{subequations}\label{eq:cf:constraintform}
\begin{align}
& \underset{\xcf \,\in\,\RN^\dimsym}{\arg\min}\;\regularization(\xcf, \xorig) \label{eq:cf:constraintform:objective} \\
& \text{ s.t. } \classifier(\xcf) = \ycf  \label{eq:cf:constraintform:constraint}
\end{align}
\end{subequations}
The authors of~\cite{counterfactualcomputationsurvey} have shown that the constrained optimization problem~\refeq{eq:cf:constraintform} can be turned (or efficiently approximated) into convex programs for many standard machine learning models including generalized linear models, quadratic discriminant analysis, nearest neighbor classifiers, etc. Since convex programs can be solved efficiently~\cite{Boyd2004}, the constrained form~\cite{counterfactualcomputationsurvey} becomes superior over the original black-box modelling~\cite{counterfactualwachter} if we have access to the underlying model $\classifier(\cdot)$.
Furthermore, we can also integrate affine pre-processings like PCA into the convex programs and therefore still compute counterfactuals in the original data space~\cite{counterfactualcomputationsurvey}.

Counterfactuals stated in its simplest form, like in~\refdef{def:counterfactual} (also called closest counterfactuals), are similar to adversarial examples, since there are no guarantees that the resulting counterfactual is plausible and feasible in the data domain. As a consequence, the absence of such constraints often leads to counterfactual explanations that are not plausible~\cite{plausiblecounterfactualsartelt,counterfactualguidedbyprototypes,face}.
To overcome this problem, several approaches~\cite{plausiblecounterfactualsartelt,counterfactualguidedbyprototypes} propose to allow only those samples that lie on the data manifold - e.g. by enforcing a lower threshold $\densitythreshold > 0$ for their probability or density. In particular, the authors of~\cite{plausiblecounterfactualsartelt} build upon the constrained form~\refeq{eq:cf:constraintform} and propose the following extension of~\refeq{eq:cf:constraintform} for computing plausible counterfactuals:
\begin{subequations}\label{eq:cf:densityconstraintform}
\begin{align}
&\underset{\xcf \,\in\,\RN^\dimsym}{\arg\min}\;\regularization(\xcf, \xorig)\label{eq:cf:densityconstraintform:objective}\\
&\quad \text{s.t. } \classifier(\xcf) = \ycf \label{eq:cf:densityconstraintform:classifierconstraint}\\
&\quad\quad\;\; \densityestimator_{\ycf}(\xcf) \geq \densitythreshold  \label{eq:cf:densityconstraintform:densityconstraint}
\end{align}
\end{subequations}
where $\densityestimator_{{\ycf}}(\cdot)$ denotes a class dependent density estimator.
Because the true density is usually not known, the density constraint  \refeq{eq:cf:densityconstraintform:densityconstraint} can be replaced with an approximation via a Gaussian mixture model (GMM) which then can be phrased as a set of convex quadratic constraints~\cite{counterfactualcomputationsurvey}.

Unstable counterfactuals can be considered as unfair to an individual in some settings, since it might happen, that counterfactual explanations are conceptually very different for two similar persons: in particular if  $\regularization(\cdot)$ is the $l_1$ norm, explanations might choose entirely different features based on which to explain the decision for neighbored samples. Next, we propose a formalization of robustness of counterfactual explanations.

\section{Robustness of Counterfactual Explanations}
We are not interested in robustness of the model itself, but the properties of the counterfactual explanations which are provided to explain the model. 
In the sub sequel we formally study and define robustness of counterfactual explanations in general, and investigate its specific properties for important specific prediction functions $\classifier:\setx\to\sety$.

\subsection{Formalization of Robustness}\label{sec:counterfactual:fairness}
We aim for a formalization of robustness of counterfactual explanations. Inspired by the intuition (and formalization) of robustness of explanation methods~\cite{DBLP:journals/corr/abs-1806-08049}), we propose the following definition for locally measuring the influence of small perturbations of the input to the counterfactual explanation:
\begin{definition}[Local instability of counterfactual explanations]\label{def:individualfairnesscontrastiveexplanation}
Let $\classifier: \setx\to\sety$ be a prediction function and $(\xorig,\yorig) \in \setx\times\sety$ with $\classifier(\xorig)=\yorig$ be a sample prediction that has to be explained. Let $\xcf = \CF{\xorig}{\ycf}\in\setx$ be a counterfactual explanations of $\xorig$.

Let $\x$ be a  perturbed sample of $\xorig$ 
that is $\x \sim \PerturbationDensity(\xorig)$ where $\PerturbationDensity(\cdot)$ denotes the density of the perturbed samples.%$\dist(\xorig,\x) \leq \epsilon$ for some suitable\todo{What does suitable mean?} metric %$\dist(\cdot)$ and
%perturbation size $\epsilon$ \todo{how is eps to be choosen?}.
%We denote the according ball of all possible perturbations as $\setB_{\epsilon}(\xorig)$.
 Denote $y=y'$ if $\classifier(\x)=\yorig$ and $y=\yorig$ otherwise.
Let $\xcfNew = \CF{\x}{y}\in\setx$ be a counterfactual explanations of this perturbed sample $(\x,\yorig)$.

Given a function $\dist(\cdot)$ for computing the similarity/distance between two given counterfactual explanations, we measure the amount of stability of the explanation $\xcf$ as the expected similarity/distance between the counterfactual explanations of the original sample $\xorig$ and such a perturbed sample $\x$:
\begin{equation}\label{eq:individualfairnesscontrastiveexplanation:score}
    \underset{\x\, \sim\, \PerturbationDensity(\xorig)}{\E}\big[\dist(\xcf, \xcfNew)\big] 
\end{equation}%\todo{too complicated, use integral notation! What is the underlying measure? Why is this well defined?}\todo{Think this only makes sense for eps -> 0, but then it is just Lip(CF), not worth noting!}
assuming that~\refeq{eq:individualfairnesscontrastiveexplanation:score} exists and is well defined.%TODO
\end{definition}
%\begin{remark}
%Monte Carlo simulations, i.e.\ replacing the expectation in~\refeq{eq:individualfairnesscontrastiveexplanation:score} with a sample mean gives us a method for empirically comparing the stability (according to \refdef{def:individualfairnesscontrastiveexplanation}) of different counterfactual explanations. \todo{No, only under further assumptions. MC does not need to converge here}
%\end{remark}
Note that the instability~\refeq{eq:individualfairnesscontrastiveexplanation:score} is to be minimized - i.e. smaller values correspond to a better robustness.

\subsection{Natural Perturbations}\label{sec:perturbations}%Why these perturbations?
While there are different  possible ways of perturbing a given input $\xorig$ - i.e.\ choosing $\PerturbationDensity(\cdot)$ in~\refdef{def:individualfairnesscontrastiveexplanation} -, we focus on three specific perturbations in this work: Perturbation by Gaussian noise, bounded uniform noise and a perturbation by masking features.

\subsubsection{Gaussian Noise}
Perturbing a given input $\xorig$ with Gaussian noise means to add a small amount of normally distributed noise  $\vec{\delta}$ to $\xorig$:
\begin{equation}\label{eq:perturbation:gaussiannoise}
    \x = \xorig + \vec{\delta} \quad \text{where } \vec{\delta} \sim \N(\vec{0}, \CovMat)
\end{equation}
we therefore define:
\begin{equation}
    \PerturbationDensity(\xorig) = \N(\xorig, \CovMat)
\end{equation}
%\todo{\x = \x ? new macro for perturbed \x}
where the size and shape of the perturbation can be controlled by the covariance matrix $\CovMat$ - in this work we use a diagonal matrix and often choose $\CovMat=\I$. However, note that in this perturbation we cannot guarantee that the perturbed sample stays close to the original sample $\xorig$ because $\N(\vec{0}, \CovMat)$ can yield arbitrarily large values. The probability of such values is limited, however. 

\subsubsection{Bounded Uniform Noise}
Similar to a perturbation with Gaussian noise (see~\refeq{eq:perturbation:gaussiannoise}), we add a small amount of uniformly distributed noise $\vec{\delta}$ to $\xorig$:
\begin{equation}\label{eq:perturbation:boundeduniformnoise}
 \x = \xorig + \vec{\delta} \quad \text{where } \vec{\delta} \sim \U(-\epsilon\vec{1}, \epsilon\vec{1})
\end{equation}
we therefore define:
\begin{equation}
    \PerturbationDensity(\xorig) = \U(-\epsilon\xorig, \epsilon\xorig)
\end{equation}
where $\epsilon > 0 $ controls and upper bounds the amount of noise - note that in~\refeq{eq:perturbation:boundeduniformnoise} we use the same $\epsilon$ for every dimension, however using different $\epsilon$ for each dimension is also possible.

Note that, unlike a perturbation with Gaussian noise, we can guarantee that the perturbed sample stays close to the original sample.
%\TODO{only in $||\cdot||_\infty$, in L2, it is $\sqrt{d}\epsilon$ close, L1 then $d\epsilon$}

\subsubsection{Feature Masking}
While a perturbation by Gaussian noise~\refeq{eq:perturbation:gaussiannoise} or bounded uniform noise~\refeq{eq:perturbation:boundeduniformnoise}\footnote{Although one could use different (possible empty) intervals for different dimensions.} potentially changes every feature, feature masking allows a  more precise way of perturbing a given input:
\begin{equation}\label{eq:perturbation:featuremasking}
    \x = \xorig \odot \mask \quad \text{where } \mask \in \{0, 1\}^\dimsym
\end{equation}
where $\odot$ denotes the element wise multiplication, and the size of the perturbation can be controlled by the number of masked features. Also note that the number of $0$s (number of masked features) as well as their position (feature id) can vary - in this work, due to the absence of expert knowledge, given a fixed number of masked features, we select the masked features randomly.
Note that in comparison to Gaussian and uniform noise, this is not a proper density and how close the perturbed sample stays to the original sample completely depends on the chosen mask.
%\TODO{so, you only put some features to 0 and all other entries are fixed? That's totally not not close to the sample anymore and rely on the entry in that feature}

\subsection{Analyzing Robustness}\label{sec:counterfactuals:robustness}
In this section we formally study robustness of closest counterfactuals under different prediction functions $\classifier(\cdot)$.
First, we give a (rather trivial) general bound on the robustness of closest counterfactual explanations in~\reftheorem{theorem:boundpeturbedcounterfactual:general} and then study each type of perturbation separately.
\begin{theorem}[General bound on closest counterfactuals of perturbed samples]\label{theorem:boundpeturbedcounterfactual:general}
Let $\classifier:\setx\to\sety$ be a prediction function and let $(\xorig,\yorig)\in\setx\times\sety$ be a sample for which we are given a closest counterfactual (see~\refeq{eq:cf:constraintform}) $\xcf=\CF{\xorig}{\ycf}\in\setx$. Let $\x\in\setx$ be a perturbed version of $\xorig$ such that $\pnorm{\xorig - \x}_\text{p} \leq \epsilon$ -- we denote the corresponding closest counterfactual of $\x$ 
%with the same target prediction $\ycf$ 
as $\xcfNew$.
We can then bound the difference between the two counterfactuals $\xcf$ and $\xcfNew$ as follows:
\begin{equation}\label{eq:boundpeturbedcounterfactual:general}
    \pnorm{\xcf - \xcfNew}_\text{p} \leq 2\epsilon + 2 \pnorm{\xorig - \xcf}_\text{p}
\end{equation}
\end{theorem}%So etwas in der Art kann man auch für pertinent positives machen :)
%\todo{Furthermore is eps+2d not suffices? }
In case of a binary linear classifier, we can refine the bound from~\reftheorem{theorem:boundpeturbedcounterfactual:general} as stated in~\refcorollary{corollary:boundpeturbedcounterfactual:binarylinearmodel}.
\begin{corollary}[Bound on closest counterfactuals of perturbed samples for binary linear classifiers]\label{corollary:boundpeturbedcounterfactual:binarylinearmodel}
In case of a binary linear classifier, i.e. $\classifier(\x)=\sign\left(\w^\top\x\right)$ with 
$\pnorm{\w}_\text{p}=1$, we can refine the bound from~\reftheorem{theorem:boundpeturbedcounterfactual:general} as follows:
\begin{equation}\label{eq:boundpeturbedcounterfactual:binarylinearmodel}
    \pnorm{\xcf - \xcfNew}_\text{p} \leq 2\epsilon + 2|\w^\top\xorig|
\end{equation}
\end{corollary}%TODO: Projektion auf Decision boundary problematisch?
\begin{remark}
Note that while the general bound~\refeq{eq:boundpeturbedcounterfactual:general} in~\reftheorem{theorem:boundpeturbedcounterfactual:general} depends on the closest counterfactual of the unperturbed sample $\xorig$, the bound~\refeq{eq:boundpeturbedcounterfactual:binarylinearmodel} in~\refcorollary{corollary:boundpeturbedcounterfactual:binarylinearmodel} only depends on the original sample $\xorig$, the model parameter $\w$ and the perturbation bound $\epsilon$.
Both bounds (\reftheorem{theorem:boundpeturbedcounterfactual:general} and~\refcorollary{corollary:boundpeturbedcounterfactual:binarylinearmodel}) are rather loose because they do not make any assumption on the perturbation $\PerturbationDensity(\cdot)$ except that it must be bounded by $\epsilon$ - in addition the bound in~\reftheorem{theorem:boundpeturbedcounterfactual:general} does not even make any assumption on $\classifier(\cdot)$ at all.
\end{remark}

\subsubsection{Gaussian Noise}
Additional assumptions allow more precise (and potentially more useful) statements as shown in the next \reftheorem{theorem:individualfairness:gaussiannoise:linearbinaryclassifier}.
There (and the consequential \refcorollary{corollary:individualfairness:gaussiannoise:linearbinaryclassifier1} and~\refcorollary{corollary:individualfairness:gaussiannoise:linearbinaryclassifier2}) we study the robustness of closest counterfactual explanations of a binary linear classifier under Gaussian noise. It turns out, that the robustness
as defined in~\refdef{def:individualfairnesscontrastiveexplanation} 
of closest counterfactual explanations of a binary linear classifier under Gaussian noise depends on the dimension $\dimsym$ only -- i.e.\ the larger the dimension of the input space, the larger the instability.%So etwas wie "curse of dimensionality for individual fairness of closest counterfactuals" ^^
\begin{theorem}[Instability of closest counterfactuals of a linear binary classifier under Gaussian noise]\label{theorem:individualfairness:gaussiannoise:linearbinaryclassifier}
Let $\classifier:\RN^\dimsym\to\{-1, 1\}$ be a binary linear classifier -- i.e. $\classifier(\x)=\sign\left(\w^\top\x\right)$ with $\pnorm{\w}_{2}=1$. The instability of closest counterfactuals (see~\refdef{def:individualfairnesscontrastiveexplanation}) under Gaussian noise~\refeq{eq:perturbation:gaussiannoise} - with an arbitrary diagonal covariance $\CovMat=\diag(\sigma_i^2)$ - at a sample $(\xorig,\yorig)\in\RN^\dimsym\times\{-1,1\}$ can be stated as follows:
\begin{equation}\label{eq:individualfairness:gaussiannoise:linearbinaryclassifier}
    \underset{\x \,\sim\, \N(\xorig,\CovMat)}{\E}\big[\dist(\xcf, \xcfNew)\big] = \trace(\CovMat) -\w^\top\CovMat\w
\end{equation}
where we assume the squared Euclidean distance as a distance $\dist(\cdot)$ for measuring the distance between two counterfactuals.
\end{theorem}
\begin{corollary}\label{corollary:individualfairness:gaussiannoise:linearbinaryclassifier1}
If we assume the identity matrix $\I$ as a covariance matrix $\CovMat$ of the Gaussian noise in~\reftheorem{theorem:individualfairness:gaussiannoise:linearbinaryclassifier},~\refeq{eq:individualfairness:gaussiannoise:linearbinaryclassifier} simplifies as follows:
\begin{equation}\label{eq:individualfairness:gaussiannoise:linearbinaryclassifier1}
    \underset{\x \,\sim\, \N(\xorig,\I)}{\E}\big[\dist(\xcf, \xcfNew)\big] = \dimsym - 1
\end{equation}
\end{corollary}
\begin{corollary}\label{corollary:individualfairness:gaussiannoise:linearbinaryclassifier2}
In the setting of \reftheorem{theorem:individualfairness:gaussiannoise:linearbinaryclassifier} and~\refcorollary{corollary:individualfairness:gaussiannoise:linearbinaryclassifier1}, the probability that the difference is larger than a given $\delta > 0$ is bounded as follows:
\begin{equation}\label{eq:individualfairness:gaussiannoise:linearbinaryclassifier2}
    \prob\Big(\dist(\xcf, \xcfNew) \geq \delta\Big) \leq \frac{\dimsym - 1}{\delta}
\end{equation}
\end{corollary}
\begin{remark}\label{remark:curseofdimensionality}
We can interpret~\reftheorem{theorem:individualfairness:gaussiannoise:linearbinaryclassifier} (and in particular the consequential \refcorollary{corollary:individualfairness:gaussiannoise:linearbinaryclassifier1} and~\refcorollary{corollary:individualfairness:gaussiannoise:linearbinaryclassifier2}) as a ``curse of dimensionality'' for robustness of closest counterfactual explanations: the larger the dimension $\dimsym$ of the data space, the larger the potential instability if the data manifold is intrinsically locally high dimensional.
\end{remark}

\subsubsection{Bounded Uniform Noise}
We can still make some statements on the robustness~\refdef{def:individualfairnesscontrastiveexplanation} of closest counterfactual explanations, when using bounded uniform noise instead of Gaussian noise, as stated in~\reftheorem{theorem:individualfairness:uniformnoise:linearbinaryclassifier} and~\refcorollary{corollary:individualfairness:uniformnoise:linearbinaryclassifier}.
\begin{theorem}[Instability of closest counterfactuals of a linear binary classifier under bounded uniform noise]\label{theorem:individualfairness:uniformnoise:linearbinaryclassifier}
Let $\classifier:\RN^\dimsym\to\{-1, 1\}$ be a binary linear classifier - i.e. $\classifier(\x)=\sign\left(\w^\top\x\right)$ with $\pnorm{\w}_{2}=1$. The instability of closest counterfactuals (see~\refdef{def:individualfairnesscontrastiveexplanation}) under a bounded uniform noise~\refeq{eq:perturbation:boundeduniformnoise} at an arbitrary  sample $(\xorig,\yorig)\in\RN^\dimsym\times\{-1,1\}$ can be stated as follows:
\begin{equation}
\underset{\x \,\sim\, \U(\xorig \pm \epsilon\vec{1})}{\E}\big[\dist(\xcf, \xcf')\big] =\frac{\epsilon^2(\dimsym-1)}{3}
\end{equation}
where we assume the squared Euclidean distance as a distance $\dist(\cdot)$ for measuring the distance between two counterfactuals.
\end{theorem}
\begin{corollary}\label{corollary:individualfairness:uniformnoise:linearbinaryclassifier}
In the setting of \reftheorem{theorem:individualfairness:uniformnoise:linearbinaryclassifier}, the probability that the instability is larger than some $\delta > 0$ can be upper bounded as:
\begin{equation}
\prob\Big(\dist(\xcf, \xcf') \geq \delta\Big) \leq \frac{\delta\epsilon^2(\dimsym - 1)}{3}
\end{equation}
\end{corollary}

\subsection{Discussion}
We will empirically confirm these mathematical findings of a potential presence of instability even for simple models in the experiments (see Section~\ref{sec:experiments}). As a mediation, we propose to add a regularization for improving the robustness~\refdef{def:individualfairnesscontrastiveexplanation} of counterfactual explanations. One solution can exist in the reference to plausible counterfactuals instead of closest ones.
One reason of instability of closest counterfactuals comes from the fact that counterfactuals can populate all dimensions such that differences can accumulate in high dimensions. This can be avoided if data are regularized according to given observations. In addition, if a nonlinear rather than linear decision boundary is present, peculiarities of the decision boundary have less effect on the
location of the counterfactuals this way.
%in case of a ``wiggly'' decision boundary small perturbations of the input cause a completely different closest counterfactual (similar to adversarial attacks). Under the assumption that the set of plausible counterfactuals is less ``wiggly'' we would expect to observe better individual fairness when considering plausible instead of closest counterfactuals. We empirically evaluate this hypothesis in Section~\ref{sec:experiments}.%TODO: Hier schauen ob man auch noch eine Schranke finden kann die sagt, dass plausible Counterfactuals "besser" sind als closest Counterfactuals.

\subsection{Relation to Individual Fairness}\label{sec:individualfairness}
Recall that, while many fairness criterions are concerned with protected groups, individual fairness focuses on individuals and requires to ``treat similar individuals similarly''~\cite{FairnessThroughAwareness}. Given a prediction function $\classifier: \setx \to \sety$, we can formalize individual fairness as follows:
\begin{equation}\label{eq:individualfairness}%TODO: Vielleicht besser auch als Definition?
    \dist(\x_1,\x_2) \leq \epsilon_1 \implies \Delta\big(\classifier(\x_1),\classifier(\x_2)\big) \leq \epsilon_2 \quad\quad \forall\,\x_1,\x_2 \in \setx
\end{equation}
where $\dist: \setx \times \setx \to \RN_{+}$ denotes a similarity measure on the individuals in $\setx$, $\epsilon_1,\epsilon_2 > 0$ denotes a threshold up to which we consider two individuals / the predictions as similar and $\Delta:\sety\times\sety \to \RN_{+}$ denotes a similarity measure on the predictions of the ML system $\classifier(\cdot)$.
A critical choice, which highly depends on the specific use-case, are the similarity measures $\dist(\cdot)$ and $\Delta(\cdot)$.
A common default choice is a p-norm assuming real-vector spaces like we do in this work.%\todo{doen't make sense to me for $\Delta$ if we are looking at classes (class 1 and 6 should be treated the same as class 1 and 2)}

As already mentioned in the introduction of this work, robustness and (individual) fairness are highly related to each other. In particular, we can interpret our definition of local instability of counterfactual explanation (\refdef{def:individualfairnesscontrastiveexplanation}) as a measure for individual fairness of counterfactual explanations - i.e. the larger the instability, the larger the individual unfairness of the counterfactual explanation. Therefore, our robustness studies can also be interpreted as a study of individual fairness.
Note that in contrast to the definition of individually fair causal recourse as proposed in~\cite{vonkuegelgen2021fairness}, our definition (\refdef{def:individualfairnesscontrastiveexplanation}) does not rely on the existence of a structural causal model, but it is formulated as a statistical criterion on the explanations (represented as samples).
%- i.e.\ our definition defines fairness locally at a specific point under a given (probabilistic) perturbation.

%\TODO{Why are you introducing Monte Carlo simulations and did not use it here? Where are you looking at plausible counterfact or do I miss a point here?}

\section{Experiments}\label{sec:experiments}
\begin{table*}[t]
\centering
\caption{Comparing the \emph{median} $l_1$ distance between counterfactual of original sample and perturbed sample (using Gaussian noise~\refeq{eq:perturbation:gaussiannoise}) - closest and plausible counterfactual explanations. Smaller values are better - best values are \textbf{highlighted}.}
\label{table:experimentresults}  
\begin{tabular}{|c||c|c||c|c||c|c|}
 \hline
 \multicolumn{1}{|c||}{\textit{Data set}} & \multicolumn{2}{|c||}{Wine} & \multicolumn{2}{|c||}{Breast cancer} & \multicolumn{2}{|c|}{Handwritten digits}  \\
  \hline
 \textit{ Method} & Closest & Plausible & Closest & Plausible & Closest & Plausible \\
 \hline
 Softmax			 & 10.16 & \textbf{1.87} & 24.04 & \textbf{22.48} & 53.71 & \textbf{48.78}\\ 
 Decision tree      & 9.25 & \textbf{2.42} & 24.05 & \textbf{23.11} & 56.56 & \textbf{49.40}\\ 
 GLVQ               & 9.95 & \textbf{1.74} & 23.34 & \textbf{21.42} & 57.66 & \textbf{49.46}\\ 
 \hline
\end{tabular}
\end{table*}
We empirically evaluate the robustness (\refdef{def:individualfairnesscontrastiveexplanation}) of closest and plausible counterfactual explanations. Furthermore, we also empirically confirm the occurrence of the ``curse of dimensionality'' (see Remark~\ref{remark:curseofdimensionality}) in case of non-linear classifiers. For this purpose, we compute closest and plausible counterfactuals of perturbed data points for a diverse set of classifiers and data sets.

\paragraph{Data sets}
We use an artificial toy data set and three standard data sets:
\begin{itemize}
\item Toy data set: A binary classification problem where each class is characterized by an isotropic Gaussian blob. The two blobs are slightly overlapping and the number of dimensions is varied - we use this data set for testing for the ``curse of dimensionality'' (see Remark~\ref{remark:curseofdimensionality}).
\item The ``Breast Cancer Wisconsin (Diagnostic) Data Set''~\cite{breastcancer} whereby we add a PCA dimensionality reduction to $5$ dimensions to the model.
\item The ``Wine data set''~\cite{winedata}.
\item The ``Optical Recognition of Handwritten Digits Data Set''~\cite{ocr} whereby we add a PCA dimensionality reduction to $40$ dimensions to the model.
\end{itemize}

\paragraph{Models}
We use the following diverse set of models: softmax regression, generalized learning vector quantization (GLVQ) and decision tree classifier.
We use the same hyperparameters across all data sets - for all vector quantization models we use $3$ prototypes per class and for all decision trees we set the maximum depth of each tree to $7$.

\begin{figure}[t]
  \caption{Gaussian blobs: \emph{Median} $l_1$ distance between counterfactual of original sample and perturbed sample (using Gaussian noise~\refeq{eq:perturbation:gaussiannoise} with $\CovMat=\I$). Smaller values are better.}
  \label{fig:exp:perturbation:curseofdimensionality}
 \begin{minipage}[b]{0.24\textwidth}
    \includegraphics[width=\textwidth]{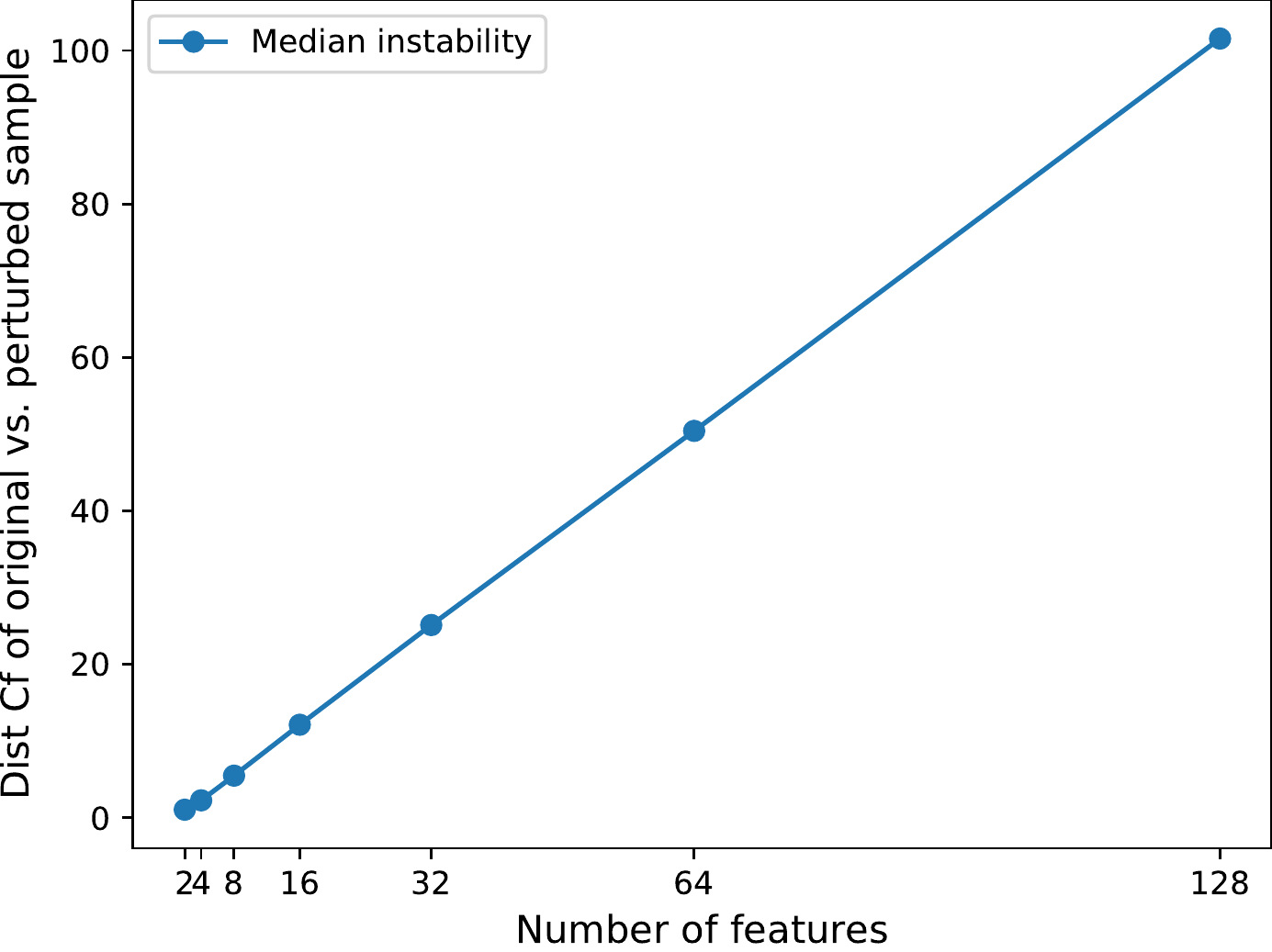}  
    \caption*{Decision tree} 
   \end{minipage}
  \hfill
  \begin{minipage}[b]{0.24\textwidth}
    \includegraphics[width=\textwidth]{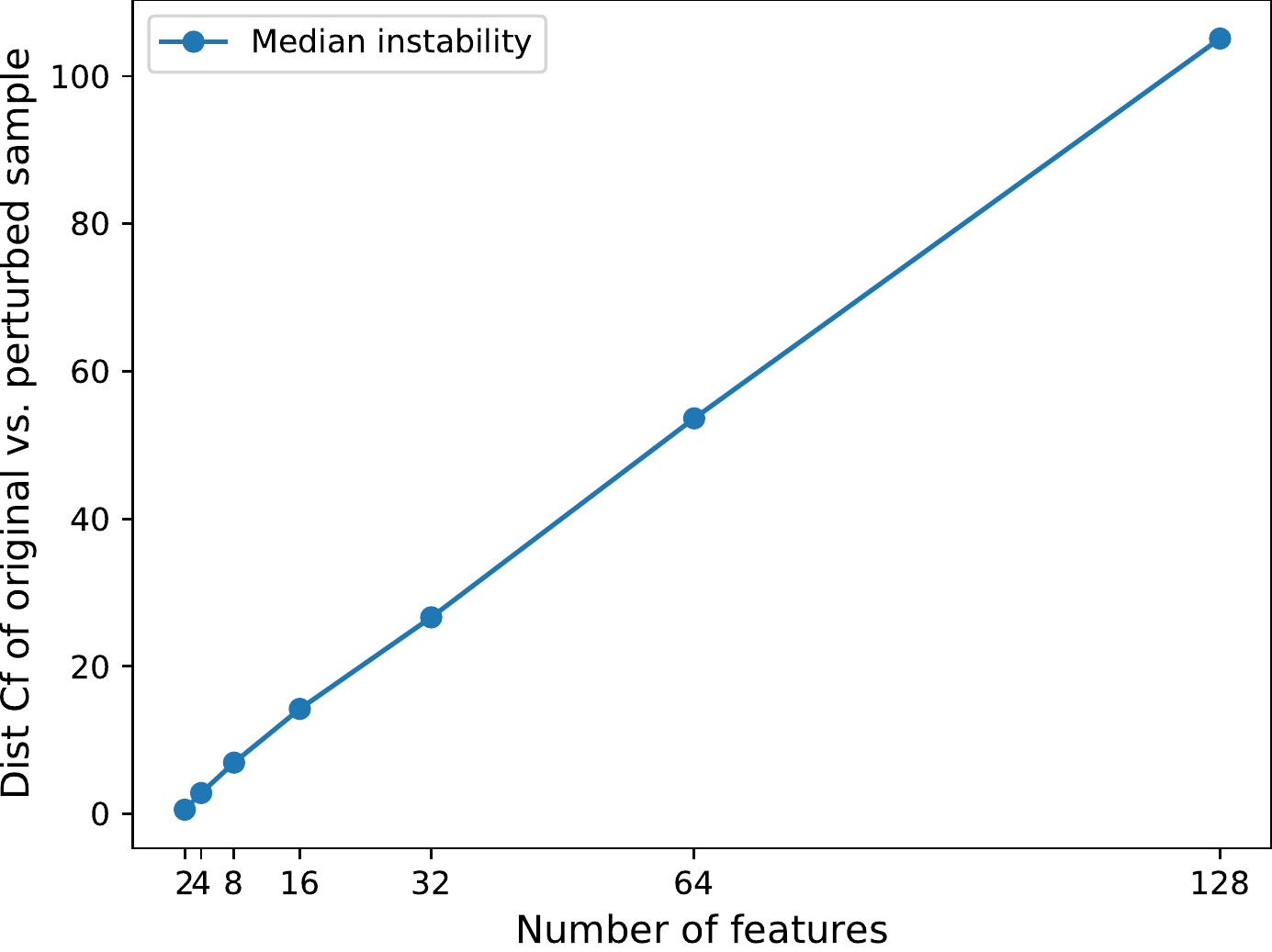}
    \caption*{GLVQ} 
 \end{minipage}
\end{figure}

\paragraph{Curse of Dimensionality}
In order to empirically study the occurrence of the ``curse of dimensionality'' (see Remark~\ref{remark:curseofdimensionality}), we use the Gaussian blobs toy data set for fitting and evaluating counterfactuals (original vs. perturbed sample) under a decision tree classifier and a GLVQ model. The results (over a $4$-fold cross validation) are shown in Fig.~\ref{fig:exp:perturbation:curseofdimensionality}.
Similar to the case of a linear classifier (see \reftheorem{theorem:individualfairness:gaussiannoise:linearbinaryclassifier}), we observe that even for non-linear classifiers, increasing the number of dimensions leads to an increase of instability.

\paragraph{Setup - Closest vs. Plausible Counterfactuals}
We report the results of the following experiments over a $4$-fold cross validation:
We fit all models on the training data (depending on the data set this might involve a PCA as a preprocessing) and compute a closest and plausible counterfactual explanations of all samples from the test set that are classified correctly by the model - whereby we compute counterfactuals of the original as well as the perturbed sample. We use two different types of perturbations:
Gaussian noise~\refeq{eq:perturbation:gaussiannoise} with $\CovMat=\I$ and
feature masking~\refeq{eq:perturbation:featuremasking} for one up to half of the total number of features.
In case of a multi-class problem, we chose a random target label that is different from the original label.
We compute and report the distance between the counterfactuals of the original sample and the perturbed sample~\refeq{eq:individualfairnesscontrastiveexplanation:score} - we do this separately for closest and plausible counterfactuals.
Furthermore, we use MOSEK\footnote{We gratefully acknowledge an academic license provided by MOSEK ApS.} as a solver for all mathematical programs. The complete implementation of the experiments is available on GitHub\footnote{\url{https://github.com/andreArtelt/FairnessRobustnessContrastingExplanations}}.

\paragraph{Results}
The results of using Gaussian noise~\refeq{eq:perturbation:gaussiannoise} for perturbing the samples are shown in Table~\ref{table:experimentresults}. The results on the digit data set for increasingly masking more and more features~\refeq{eq:perturbation:featuremasking} are shown in Fig.~\ref{fig:exp:perturbation:featuremasking:digits} - plots for the other data sets are given in appendix~\ref{sec:appendix:experiments:plots}.

We observe that in all cases the plausible counterfactual explanations are less affected by perturbations than the closest counterfactuals - thus we consider them to be more robust (\refdef{def:individualfairnesscontrastiveexplanation}). The size of the differences depends a lot on the combination of model and data set. However, in all cases there is a clear difference. %TODO: Eigentlich Statistik mit HypothesenTest notwendig xD
In case of increasingly masking features, we observe that although the distance between counterfactuals of original and perturbed sample is subject to some variance, the plausible counterfactuals are more robust than the closest counterfactuals - even when masking up to $50$\% of all features.

\section{Discussion and Conclusion}\label{sec:conclusion}
\begin{figure*}[t]
  \caption{Handwritten digits data set: \emph{Median} $l_1$ distance between counterfactual of original sample and perturbed sample (using feature masking~\refeq{eq:perturbation:featuremasking}) for closest and plausible counterfactual explanations - for different number of masked features. Smaller values are better.}
  \label{fig:exp:perturbation:featuremasking:digits}
 \begin{minipage}[b]{0.3\textwidth}
    \includegraphics[width=\textwidth]{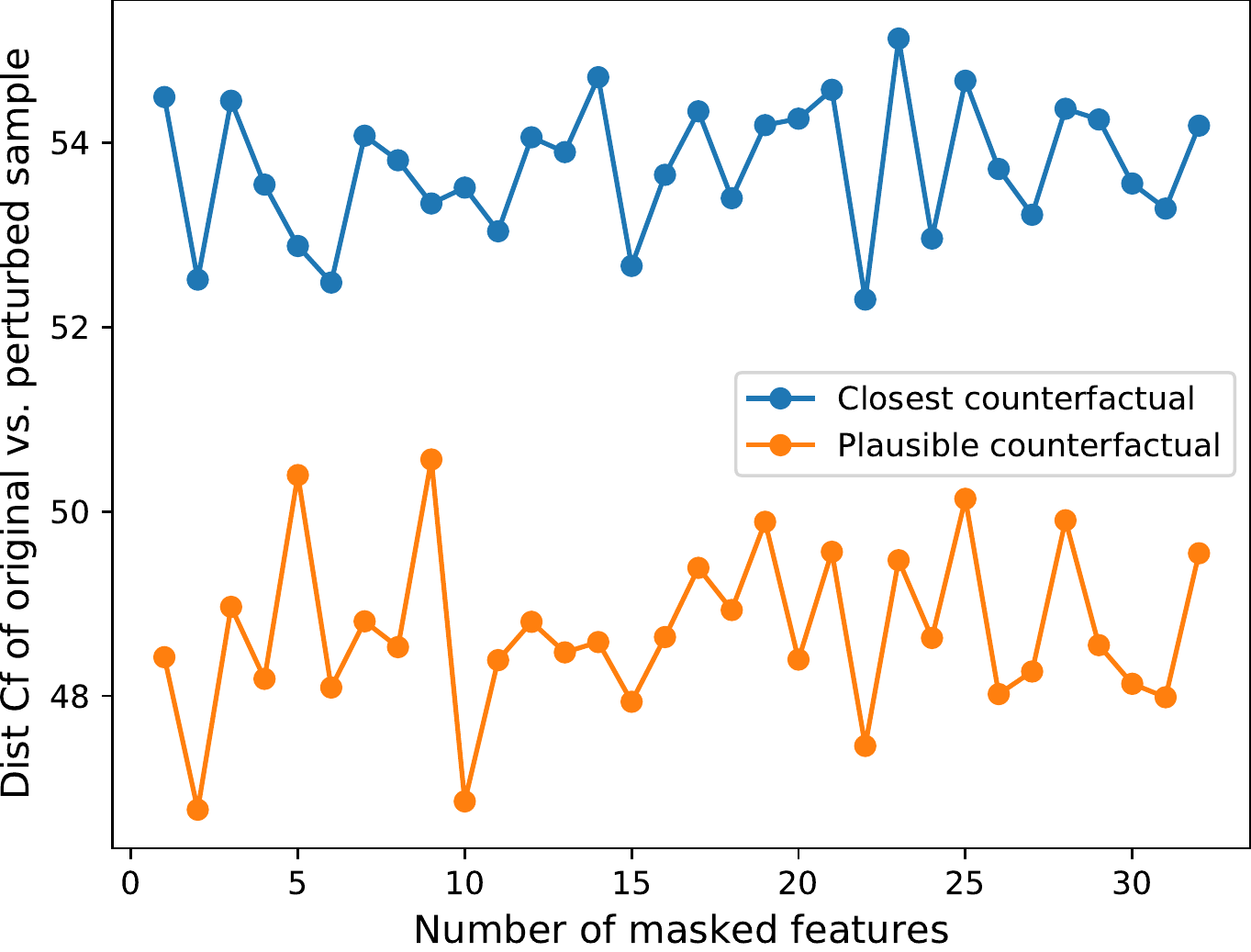}  
    \caption*{Softmax regression} 
   \end{minipage}
  \hfill
  \begin{minipage}[b]{0.3\textwidth}
    \includegraphics[width=\textwidth]{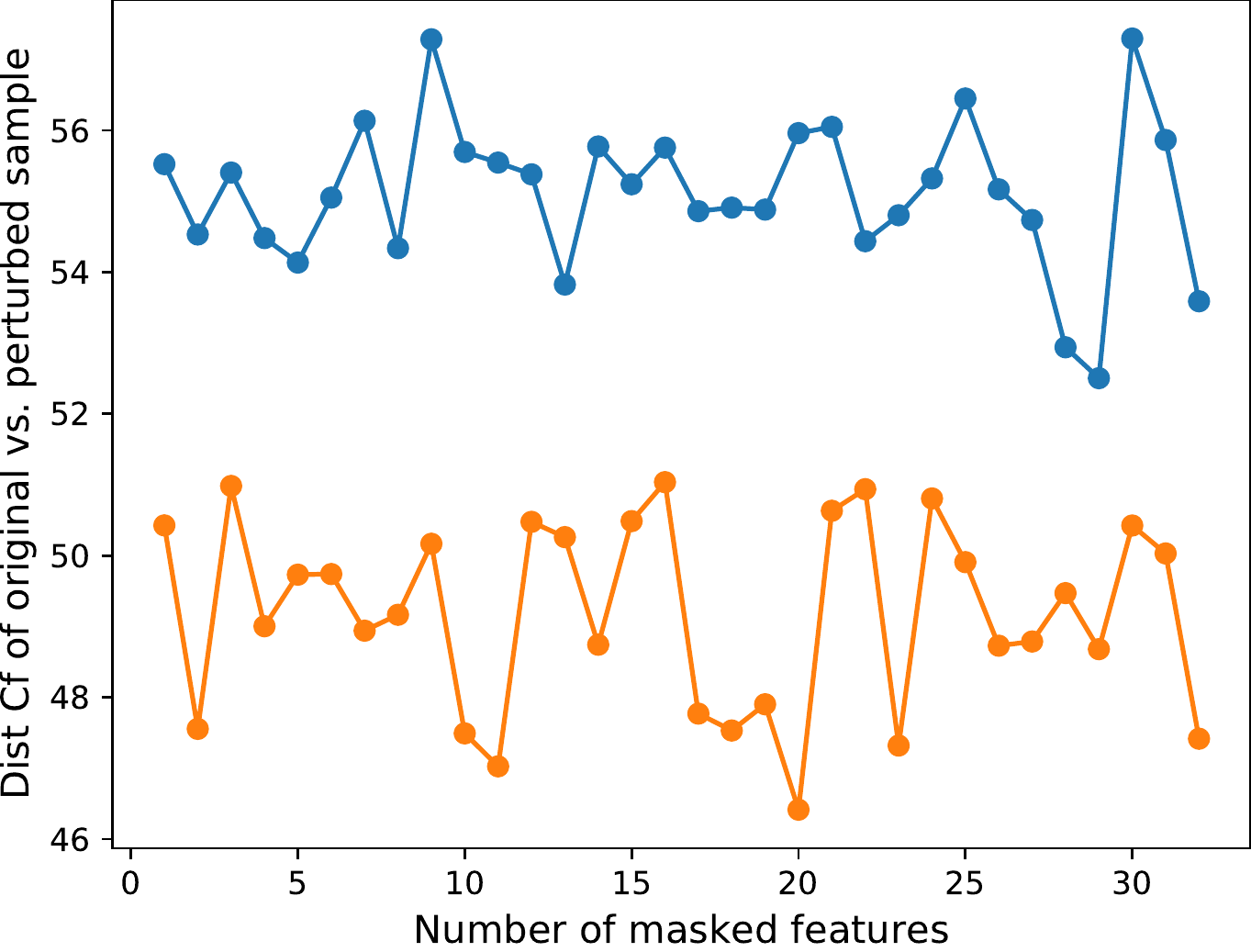}
    \caption*{Decision tree} 
  \end{minipage}
  \hfill
  \begin{minipage}[b]{0.3\textwidth}
    \includegraphics[width=\textwidth]{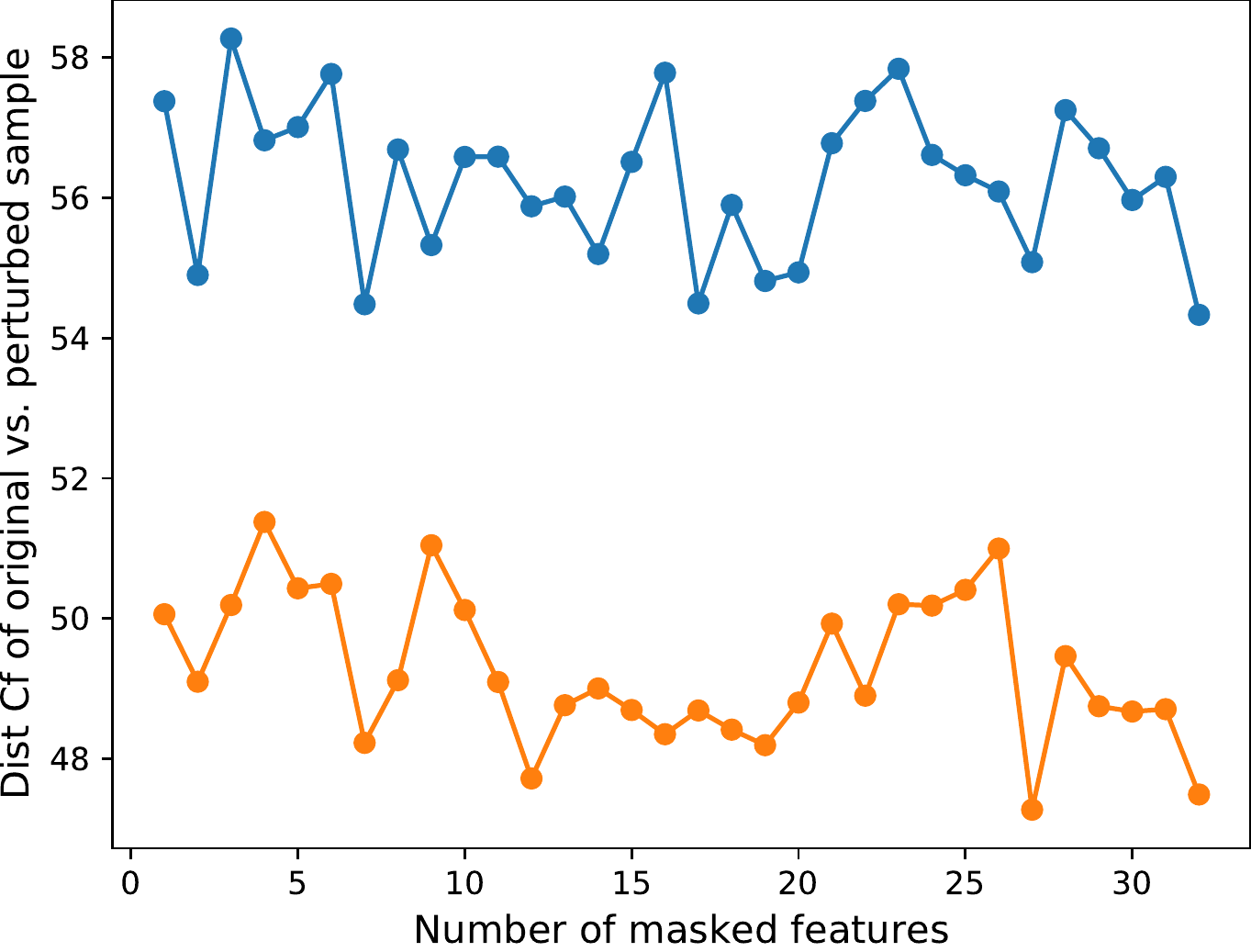}
    \caption*{GLVQ} 
 \end{minipage}
\end{figure*}
%\todo{dist vs dist? That's not what you are plotting here on the y-axis!}
In this work, we studied the robustness of counterfactual explanations. We found that closest counterfactuals can be unstable in the sense that they are sensitive to different kinds of small perturbations. We proposed to use plausible instead of closest counterfactuals for increasing the robustness of counterfactual explanations which we empirically evaluated providing a  comparison of the robustness of closest vs. plausible counterfactual explanations. We found evidence that plausible counterfactuals provide better robustness than closest counterfactual explanations.
We also argued that robustness and individual fairness of counterfactual explanations are basically the same and thus, our findings on robustness also apply to individual fairness of counterfactual explanations --- i.e. the individual fairness of closest counterfactual explanation is rather poor and using plausible counterfactuals yield a better individual fairness.

In future work, we plan to further study formal fairness and robustness guarantees and bounds of more models (e.g. deep neural networks) and different perturbations. We also would like to investigate other approaches and methodologies for computing plausible counterfactual explanations---the work~\cite{plausiblecounterfactualsartelt} we used in this article is only one possible approach for computing plausible counterfactuals, for other approaches see~\cite{face,counterfactualguidedbyprototypes}. Finally, we are highly interested in studying the problem of individual fairness of (contrasting) explanations from a psychological perspective, i.e. investigating how people actually experience individual fairness of contrasting explanations and whether this experience is successfully captured/modeled by our proposed formalization and methods.

%\section*{References}

\bibliographystyle{IEEEtran}
\bibliography{bibliography}

\appendix
\section{Proofs and Derivations}\label{sec:appendix}
\begin{enumerate}
\item
\begin{IEEEproof}[Theorem~\ref{theorem:boundpeturbedcounterfactual:general}]
Since the perturbation is bounded by $\epsilon>0$, it holds that:
\begin{equation}\label{eq:boundpeturbedcounterfactual:general:boundednoise}
    \pnorm{\xorig-\x}_\text{p} \leq \epsilon
\end{equation}
Furthermore, if the closest counterfactual $\xcf$ of $\xorig$ is different from the closest counterfactual $\xcfNew$ of $\x$ (perturbed $\xorig$), it must hold that:
\begin{equation}\label{eq:boundpeturbedcounterfactual:general:cfdifference}
    \pnorm{\x-\xcfNew}_\text{p} \leq \pnorm{\x-\xcf}_\text{p}
\end{equation}
Because of the triangle inequality we know that the following holds:
\begin{equation}\label{eq:boundpeturbedcounterfactual:general:cfdifference2}
    \pnorm{\xcfNew - \xcf}_\text{p} \leq \pnorm{\x - \xcfNew}_\text{p} + \pnorm{\x - \xcf}_\text{p}
\end{equation}
Plugging~\refeq{eq:boundpeturbedcounterfactual:general:cfdifference} into~\refeq{eq:boundpeturbedcounterfactual:general:cfdifference2} yields:
\begin{equation}\label{eq:boundpeturbedcounterfactual:general:final1}
\begin{split}
    \pnorm{\xcf - \xcfNew}_\text{p} &\leq \pnorm{\x - \xcfNew}_\text{p} + \pnorm{\x - \xcf}_\text{p} \\
    &\leq \pnorm{\x-\xcf}_\text{p} + \pnorm{\x-\xcf}_\text{p} \\
    &= 2\pnorm{\x - \xcf}_\text{p}
\end{split}
\end{equation}
By making use of the triangle inequality and~\refeq{eq:boundpeturbedcounterfactual:general:boundednoise}, we find that:
\begin{equation}\label{eq:boundpeturbedcounterfactual:general:final2}
\begin{split}
    \pnorm{\x - \xcf}_\text{p} &\leq \pnorm{\xorig-\x}_\text{p} + \pnorm{\xorig - \xcf}_\text{p} \\
    &\leq \epsilon + \pnorm{\xorig - \xcf}_2
\end{split}
\end{equation}
Plugging~\refeq{eq:boundpeturbedcounterfactual:general:final2} into~\refeq{eq:boundpeturbedcounterfactual:general:final1} yields the desired bound~\refeq{eq:boundpeturbedcounterfactual:general}:
\begin{equation}
\begin{split}
    \pnorm{\xcf - \xcfNew}_\text{p} &\leq 2\pnorm{\x - \xcf}_\text{p} \\
    &= 2\epsilon + 2\pnorm{\xorig - \xcf}_\text{p}
\end{split}
\end{equation}
\end{IEEEproof}

\item
\begin{IEEEproof}[Corollary~\ref{corollary:boundpeturbedcounterfactual:binarylinearmodel}]
First, we prove that the closest counterfactual explanations $\xcf$ of a sample $\xorig$ under a binary linear classifier $\classifier(\x)=\sign(\w^\top\x)$ (we assume w.l.o.g. $\pnorm{\w}_2=1$) can be explicitly stated as follows:
\begin{equation}\label{eq:closestcf:binarylinearclassifier}
    \xcf = \xorig - \left(\w^\top\xorig\right)\w
\end{equation}%TODO: Könnte man als separates Lemma machen.
Computing the closest counterfactual of some $\xorig$ under a binary linear classifier can be formalized as the following optimization problem:
\begin{subequations}\label{eq:binarylinearmodel:closestcf:optproblem}
\begin{align}
&\underset{\xcf\,\in\,\RN^\dimsym}{\min}\, \pnorm{\xorig - \xcf}_2^2 \label{eq:binarylinearmodel:closestcf:optproblem:objective}\\
&\text{s.t. } \w^\top\xcf = 0 \label{eq:binarylinearmodel:closestcf:optproblem:constraint}
\end{align}
\end{subequations}
Note that the constraint~\refeq{eq:binarylinearmodel:closestcf:optproblem:constraint} ``replaces/approximates'' the constraint $\classifier(\xcf)=\ycf$~\refeq{eq:cf:constraintform:constraint}. The constraint~\refeq{eq:binarylinearmodel:closestcf:optproblem:constraint} requires that the solution $\xcf$ lies directly on the decision boundary. We assume that points on the decision boundary are classified as $\ycf$ - while this approach is debatable, it offers an easy solution to the original problem because otherwise we would have to project onto an open set which is ``difficult'' (once we are on the decision boundary we could add an infinitesimally small constant to the solution for crossing the decision boundary if this is really necessary). % Etwas "geschummelt", aber wie im Text gesagt wird, geht es halt nichts anders. Notfalls muss man halt noch ein kleines Epsilon drauf addieren, wenn man unbedingt über die Entscheidungsgrenze springen möchte ;)
% Wenn man y(w^Tx) + e <= 0 als Constraint nimmt, bekommt man einen zusätzlichen Term x = ... - y*e*w  Dieser zusätzliche Term verschwindet aber, wenn man lim e -> 0 (x) nimmt. Außerdem sind die Bound bzgl. der unfairness so wie so unabhängig von x. Daher kann man den auch gleich ignorieren (man könnte argumentieren, dass man einfach das Koordinatensystem shifted, so dass ein beliebiges epsilon>0 berücksichtig wird) - Problem ist halt, dass das closest counterfactual in einer offenen Menge liegt und daher nicht so gut zu bestimmen ist xD Ich denke daher, dass der Ansatz auf die Decision Boundary zo projezieren vertretbar ist.

We solve~\refeq{eq:binarylinearmodel:closestcf:optproblem} by using the method of Lagrangian multipliers. Since~\refeq{eq:binarylinearmodel:closestcf:optproblem} is a convex optimization problem, we only have globally optimal solutions - in particular we have a quadratic objective and an linear constraint, thus strong duality holds. % Starke Dualität :)
The Lagrangian of~\refeq{eq:binarylinearmodel:closestcf:optproblem} is given as follows:
\begin{equation}\label{eq:binarylinearmodel:closestcf:lagrangian}
\Lagrange(\xcf,\lambda) = \xorig^\top\xorig - 2\xorig^\top\xcf + \xcf^\top\xcf - \lambda\w^\top\xcf
\end{equation}
The gradient of the Lagrangian~\refeq{eq:binarylinearmodel:closestcf:lagrangian} with respect to $\xcf$ can be written as follows:
\begin{equation}\label{eq:binarylinearmodel:closestcf:lagrangian:gradient}
\begin{split}
    \nabla_{\xcf}\Lagrange(\xcf, \lambda) &= \nabla_{\xcf}\xorig^\top\xorig - \nabla_{\xcf}2\xorig^\top\xcf + \nabla_{\xcf}\xcf^\top\xcf -\\&\quad \nabla_{\xcf}\lambda\w^\top\xcf\\
    &= -2\xorig + 2\xcf - \lambda\w
\end{split}
\end{equation}
The optimality condition requires the gradient~\refeq{eq:binarylinearmodel:closestcf:lagrangian:gradient} being equal to zero:
\begin{equation}\label{eq:binarylinearmodel:closestcf:lagrangian:gradientsolution}
\begin{split}
    &\nabla_{\xcf}\Lagrange(\xcf, \lambda) = \vec{0} \\
    &\Leftrightarrow -2\xorig + 2\xcf - \lambda\w = 0\\
    &\Leftrightarrow \xcf = \xorig + \frac{\lambda}{2}\w
\end{split}
\end{equation}
Plugging~\refeq{eq:binarylinearmodel:closestcf:lagrangian:gradientsolution} back into the Lagrangian~\refeq{eq:binarylinearmodel:closestcf:lagrangian} yields the Lagrangian dual:
\begin{equation}\label{eq:binarylinearmodel:closestcf:lagrangian:dual}
\begin{split}
    &\Lagrange_{D}(\lambda) = \underset{{\xcf\,\in\,\RN^\dimsym}}{\min}\,\Lagrange(\xcf,\lambda)\\
    &\Lagrange\left(\xcf = \xorig + \frac{\lambda}{2}\w, \lambda\right)=\xorig^\top\xorig - 2\xorig^\top\left(\xorig + \frac{\lambda}{2}\w\right) +\\&\quad \left(\xorig + \frac{\lambda}{2}\w\right)^\top\left(\xorig + \frac{\lambda}{2}\w\right) - \lambda\w^\top\left(\xorig + \frac{\lambda}{2}\w\right) \\
    &=\xorig^\top\xorig - 2\xorig^\top\xorig - \lambda\xorig^\top\w + \xorig^\top\xorig +\\&\quad \lambda\xorig^\top\w + \frac{\lambda^2}{4}\w^\top\w - \lambda\xorig^\top\w - \frac{\lambda^2}{2}\w^\top\w \\
    &= \frac{\lambda^2}{4}\w^\top\w - \lambda\xorig^\top\w - \frac{\lambda^2}{2}\w^\top\w  \\
    &= -\frac{\lambda^2}{4}\w^\top\w - \lambda\xorig^\top\w
\end{split}
\end{equation}
The gradient of the Lagrangian dual~\refeq{eq:binarylinearmodel:closestcf:lagrangian:dual} can be written as follows:
\begin{equation}\label{eq:binarylinearmodel:closestcf:lagrangian:dual:gradient}
\begin{split}
    \frac{\partial}{\partial \lambda}\Lagrange_{D}(\lambda) &= -\frac{\partial}{\partial \lambda}\frac{\lambda^2}{4}\w^\top\w - \frac{\partial}{\partial \lambda}\lambda\xorig^\top\w \\
    &=  -\frac{\lambda}{2}\w^\top\w - \xorig^\top\w
\end{split}
\end{equation}
Next, the optimality condition requires that the gradient~\refeq{eq:binarylinearmodel:closestcf:lagrangian:dual:gradient} is equal to zero:
\begin{equation}\label{eq:binarylinearmodel:closestcf:lagrangian:dual:solution}
\begin{split}
    & \frac{\partial}{\partial \lambda}\Lagrange_{D}(\lambda) = 0\\
    &\Leftrightarrow -\frac{\lambda}{2}\w^\top\w - \xorig^\top\w = 0 \\
    &\Leftrightarrow \lambda = -2\xorig^\top\w = -2\w^\top\xorig
\end{split}
\end{equation}
where we made use of $\pnorm{\w}_2=1$.

Finally, we obtain the solution of the original problem~\refeq{eq:binarylinearmodel:closestcf:optproblem} by plugging the solution of the dual problem~\refeq{eq:binarylinearmodel:closestcf:lagrangian:dual:solution} into~\refeq{eq:binarylinearmodel:closestcf:lagrangian:gradientsolution}:
\begin{equation}
\begin{split}
    \xcf &= \xorig + \frac{\lambda}{2}\w \\
    &= \xorig + \frac{-2\w^\top\xorig}{2}\w \\
    &= \xorig - \left(\w^\top\xorig\right)\w
\end{split}
\end{equation}
which concludes this sub-proof.

Plugging~\refeq{eq:closestcf:binarylinearclassifier} into the bound~\refeq{eq:boundpeturbedcounterfactual:general} from~\reftheorem{theorem:boundpeturbedcounterfactual:general}, and again assuming w.l.o.g. that $\pnorm{\w}_2=1$, yields the desired bound~\refeq{eq:boundpeturbedcounterfactual:binarylinearmodel}:
\begin{equation}
\begin{split}
    \pnorm{\xcf - \xcfNew}_2 &\leq 2\epsilon + 2 \pnorm{\xorig - \xcf}_2 \\
    &= 2\epsilon + 2\pnorm{\xorig - (\xorig - \w^\top\xorig\w)}_2 \\
    &= 2\epsilon + 2\pnorm{\w^\top\xorig\w}_2\\
    &= 2\epsilon + 2|\w^\top\xorig|
\end{split}
\end{equation}
\end{IEEEproof}

\item
\begin{IEEEproof}[\reftheorem{theorem:individualfairness:gaussiannoise:linearbinaryclassifier}]
From the proof of~\refcorollary{corollary:boundpeturbedcounterfactual:binarylinearmodel} we now that the closest counterfactual explanation $\xcf$ of a sample $\x$ under a linear binary classifier $\classifier(\x)=\w^\top\x$ can be stated explicitly~\refeq{eq:closestcf:binarylinearclassifier}:
\begin{equation}\label{eq:individualfairness:gaussiannoise:linearbinaryclassifier:cloestformcf}
    \xcf = \x - \left(\w^\top\x\right)\w
\end{equation}
Applying the analytic solution~\refeq{eq:individualfairness:gaussiannoise:linearbinaryclassifier:cloestformcf} to the squared Euclidean distance between the closest counterfactual $\xcf$ of the original sample $\xorig$ and the closest counterfactual $\xcfNew$ of the corresponding perturbed sample $\x$ yields:
\begin{equation}\label{eq:individualfairness:gaussiannoise:linearbinaryclassifier:cfdist}
\begin{split}
    \dist(\xcf,\xcfNew) &= \left(\xcf - \xcfNew\right)^\top\left(\xcf - \xcfNew\right) \\
    &= \xcf^\top\xcf - 2\xcf^\top\xcfNew +\xcfNew^\top\xcfNew \\
    &= \left(\xorig - \left(\w^\top\xorig\right)\w\right)^\top\left(\xorig - \left(\w^\top\xorig\right)\w\right) - \\&\quad2\left(\xorig - \left(\w^\top\xorig\right)\w\right)^\top\left(\x - \left(\w^\top\x\right)\w\right) + \\&\quad\left(\x - \left(\w^\top\x\right)\w\right)^\top\left(\x - \left(\w^\top\x\right)\w\right) \\
    &= \xorig^\top\xorig - 2\xorig^\top\x - 2\left(\w^\top\xorig\right)^2
     + 2\xorig^\top\left(\w^\top\x\right)\w + \\&\quad \x^\top\x
     + 2\left(\xorig^\top\w\right)\left(\x^\top\w\right) - 2\left(\x^\top\w\right)^2
     + \left(\xorig^\top\w\right)^2 - \\&\quad 2\left(\xorig^\top\w\right)^\top\left(\x^\top\w\right)
     + \left(\x^\top\w\right)^2 \\
     &= \xorig^\top\xorig - 2\xorig^\top\x - \left(\w^\top\xorig\right)^2
     +\\&\quad 2\left(\w^\top\x\right)\left(\xorig^\top\w\right) + \x^\top\x
     - \left(\x^\top\w\right)^2
\end{split}
\end{equation}
Taking the expectation of~\refeq{eq:individualfairness:gaussiannoise:linearbinaryclassifier:cfdist} over an arbitrary density $\density(\cdot)$ yields:
\begin{equation}\label{eq:individualfairness:gaussiannoise:linearbinaryclassifier:generalexpectation}
\begin{split}
    \underset{\x \,\sim\, \density}{\E}\big[\dist(\xcf,\xcfNew)\big] &= \underset{\x \,\sim\, \density}{\E}\Big[\xorig^\top\xorig - 2\xorig^\top\x - \left(\w^\top\xorig\right)^2
     + \\&\quad 2\left(\w^\top\x\right)\left(\xorig^\top\w\right) + \x^\top\x - \left(\x^\top\w\right)^2\Big]\\
    &= \xorig^\top\xorig - \E\Big[2\xorig^\top\x\Big] - \left(\w^\top\xorig\right)^2
     + \\&\quad 2\left(\xorig^\top\w\right)\E\Big[\w^\top\x\Big] + \E\Big[\x^\top\x\Big] - \E\Big[\left(\x^\top\w\right)^2\Big]
\end{split}
\end{equation}
Working out the specific expectations from~\refeq{eq:individualfairness:gaussiannoise:linearbinaryclassifier:generalexpectation} and under a Gaussian distribution $\x\sim\N(\xorig,\CovMat)$ with $\CovMat=\diag(\sigma_i^2)$ - i.e. $(\x)_i$s are uncorrelated - yields:
\begin{equation}\label{eq:individualfairness:gaussiannoise:linearbinaryclassifier:expect:1}
\begin{split}
    \E\Big[2\xorig^\top\x\Big] &= \E\Big[2\sum_i(\xorig)_i(\x)_i\Big]\\
    &= 2\sum_i(\xorig)_i\E\big[(\x)_i\big] \\
    &= 2\sum_i(\xorig)_i(\xorig)_i \\
    &= 2\xorig^\top\xorig
\end{split}
\end{equation}
\begin{equation}\label{eq:individualfairness:gaussiannoise:linearbinaryclassifier:expect:2}
\begin{split}
    \E\Big[\w^\top\x\Big] &= \E\Big[\sum_i(\w)_i(\x)_i\Big]\\
    &= \sum_i(\w)_i\E\big[(\x)_i\big] \\
    &= \sum_i(\w)_i(\xorig)_i \\
    &= \w^\top\xorig
\end{split}
\end{equation}
\begin{equation}\label{eq:individualfairness:gaussiannoise:linearbinaryclassifier:expect:3}
\begin{split}
    \E\Big[\x^\top\x\Big] &= \E\Big[\sum_i(\x)_i^2\Big]\\
    &= \sum_i\E\big[(\x)_i^2\big]\\
    &= \sum_i\left(\E\big[(\x)_i\big]^2 + \Var\big[(\x)_i\big]\right) \\
    &= \sum_i\Big((\xorig)_i^2 + \sigma_i^2\Big) \\
    &= \xorig^\top\xorig + \trace(\CovMat)
\end{split}
\end{equation}
\begin{equation}\label{eq:individualfairness:gaussiannoise:linearbinaryclassifier:expect:4}
\begin{split}
    \E\Big[\left(\x^\top\w\right)^2\Big] &= \E\big[\x^\top\w\big]^2 + \Var\big[\x^\top\w\big]\\
    &= \left(\w^\top\xorig\right)^2 + \Var\Big[\sum_i(\w)_i(\x)_i\Big] \\
    &= \left(\w^\top\xorig\right)^2 + \sum_i\Var\big[(\w)_i(\x)_i\big] \\
    &= \left(\w^\top\xorig\right)^2 + \sum_i(\w)_i^2\sigma_i^2 \\
    &= \left(\w^\top\xorig\right)^2 + \w^\top\CovMat\w
\end{split}
\end{equation}
where we made use of the assumption that $\pnorm{\w}_2=1$.

Substituting~\refeq{eq:individualfairness:gaussiannoise:linearbinaryclassifier:expect:1},~\refeq{eq:individualfairness:gaussiannoise:linearbinaryclassifier:expect:2},~\refeq{eq:individualfairness:gaussiannoise:linearbinaryclassifier:expect:3},~\refeq{eq:individualfairness:gaussiannoise:linearbinaryclassifier:expect:4} in~\refeq{eq:individualfairness:gaussiannoise:linearbinaryclassifier:generalexpectation} yields:
\begin{equation}
\begin{split}
    \underset{\x \,\sim\, \N(\xorig,\CovMat)}{\E}\big[\dist(\xcf,\xcfNew)\big] &= \xorig^\top\xorig - \E\Big[2\xorig^\top\x\Big] - \left(\w^\top\xorig\right)^2
     + \\&\quad 2\left(\xorig^\top\w\right)\E\Big[\w^\top\x\Big] + \E\Big[\x^\top\x\Big] - \E\Big[\left(\x^\top\w\right)^2\Big]\\
    &= \xorig^\top\xorig - 2\xorig^\top\xorig - \left(\w^\top\xorig\right)^2
     + \\&\quad 2\left(\xorig^\top\w\right)^2 + \xorig^\top\xorig + \trace(\CovMat) -\\&\quad \left(\w^\top\xorig\right)^2 - \w^\top\CovMat\w \\
     &= \trace(\CovMat) - \w^\top\CovMat\w
\end{split}
\end{equation}
which concludes the proof.
\end{IEEEproof}

\item
\begin{IEEEproof}[\refcorollary{corollary:individualfairness:gaussiannoise:linearbinaryclassifier1}]
Substituting $\I$ for $\CovMat$ in~\refeq{eq:individualfairness:gaussiannoise:linearbinaryclassifier} from~\reftheorem{theorem:individualfairness:gaussiannoise:linearbinaryclassifier} yields the claimed expectation:
\begin{equation}
\begin{split}
\underset{\x \,\sim\, \N(\xorig,\CovMat)}{\E}\big[\dist(\xcf, \xcfNew)\big] &= \trace(\CovMat) - \w^\top\CovMat\w \\
&= \trace(\I) - \w^\top\I\w \\
&= \dimsym - 1
\end{split}
\end{equation}
\end{IEEEproof}

\item
\begin{IEEEproof}[\refcorollary{corollary:individualfairness:gaussiannoise:linearbinaryclassifier2}]
Plugging the expectation from~\refcorollary{corollary:individualfairness:gaussiannoise:linearbinaryclassifier1} into Markov's inequality yields the claimed bound:
\begin{equation}
\begin{split}
    \prob\Big(\dist(\xcf, \xcfNew) \geq \delta\Big) &\leq \frac{{\E}\big[\dist(\xcf,\xcfNew)\big]}{\delta} \\
    &= \frac{\dimsym - 1}{\delta}
\end{split}
\end{equation}
\end{IEEEproof}

\item
\begin{IEEEproof}[\reftheorem{theorem:individualfairness:uniformnoise:linearbinaryclassifier}]
From the proof of~\reftheorem{theorem:individualfairness:gaussiannoise:linearbinaryclassifier} we know that the exceptation over an arbitrary density $\density(\cdot)$ of the distance between the closest counterfactual of the original sample and the perturbed sample can be written as follows:
\begin{equation}\label{eq:individualfairness:uniformnoise:linearbinaryclassifier:generalexpectation}
\begin{split}
    \underset{\x \,\sim\, \density}{\E}\big[\dist(\xcf,\xcfNew)\big] &= \xorig^\top\xorig - \E\Big[2\xorig^\top\x\Big] - \left(\w^\top\xorig\right)^2
     + \\&\quad 2\left(\xorig^\top\w\right)\E\Big[\w^\top\x\Big] + \E\Big[\x^\top\x\Big] - \E\Big[\left(\x^\top\w\right)^2\Big]
\end{split}
\end{equation}
Next, working out the specific expectations from~\refeq{eq:individualfairness:uniformnoise:linearbinaryclassifier:generalexpectation} under a bounded uniform noise $\x \,\sim\, \U(\xorig \pm \epsilon\vec{1})$ - i.e. $(\x)_i$s are uncorrelated - yields:
\begin{equation}\label{eq:individualfairness:uniformnoise:linearbinaryclassifier:expect:1}
\begin{split}
    \E\Big[2\xorig^\top\x\Big] &= \E\Big[2\sum_i(\xorig)_i(\x)_i\Big]\\
    &= 2\sum_i(\xorig)_i\E\big[(\x)_i\big] \\
    &= 2\sum_i(\xorig)_i \frac{1}{2}\Big((\xorig)_i - \epsilon + (\xorig)_i + \epsilon\Big) \\
    &= 2\sum_i(\xorig)_i\Big(\xorig)_i \\
    &= 2\xorig^\top\xorig
\end{split}
\end{equation}
\begin{equation}\label{eq:individualfairness:uniformnoise:linearbinaryclassifier:expect:2}
\begin{split}
    \E\Big[\w^\top\x\Big] &= \E\Big[\sum_i(\w)_i(\x)_i\Big]\\
    &= \sum_i(\w)_i\E\big[(\x)_i\big] \\
    &= \sum_i(\w)_i(\xorig)_i \\
    &= \w^\top\xorig
\end{split}
\end{equation}
\begin{equation}\label{eq:individualfairness:uniformnoise:linearbinaryclassifier:expect:3}
\begin{split}
    \E\Big[\x^\top\x\Big] &= \E\Big[\sum_i(\x)_i^2\Big]\\
    &= \sum_i\E\big[(\x)_i^2\big]\\
    &= \sum_i\left(\E\big[(\x)_i\big]^2 + \Var\big[(\x)_i\big]\right) \\
    &= \sum_i\Big((\xorig)_i^2 + \frac{1}{12}\Big((\xorig)_i + \epsilon - (\xorig)_i + \epsilon\Big)^2\Big) \\
    &= \sum_i\Big((\xorig)_i^2 + \frac{4\epsilon^2}{12}\Big) \\
    &= \xorig^\top\xorig + \frac{\dimsym\epsilon^2}{3}
\end{split}
\end{equation}
\begin{equation}\label{eq:individualfairness:uniformnoise:linearbinaryclassifier:expect:4}
\begin{split}
    \E\Big[\left(\x^\top\w\right)^2\Big] &= \E\big[\x^\top\w\big]^2 + \Var\big[\x^\top\w\big]\\
    &= \left(\w^\top\xorig\right)^2 + \Var\Big[\sum_i(\w)_i(\x)_i\Big] \\
    &= \left(\w^\top\xorig\right)^2 + \sum_i\Var\big[(\w)_i(\x)_i\big] \\
    &= \left(\w^\top\xorig\right)^2 + \sum_i(\w)_i^2\frac{\epsilon^2}{3} \\
    &= \left(\w^\top\xorig\right)^2 + \w^\top\w\frac{\epsilon^2}{3} \\
    &= \left(\w^\top\xorig\right)^2 + \frac{\epsilon^2}{3}
\end{split}
\end{equation}
Substituting~\refeq{eq:individualfairness:uniformnoise:linearbinaryclassifier:expect:1},~\refeq{eq:individualfairness:uniformnoise:linearbinaryclassifier:expect:2},~\refeq{eq:individualfairness:uniformnoise:linearbinaryclassifier:expect:3},~\refeq{eq:individualfairness:uniformnoise:linearbinaryclassifier:expect:4} in~\refeq{eq:individualfairness:uniformnoise:linearbinaryclassifier:generalexpectation} yields:
\begin{equation}
\begin{split}
    \underset{\x\,\sim\,\U(\xorig \pm \epsilon\vec{1})}{\E}\big[\dist(\xcf,\xcfNew)\big] &= \xorig^\top\xorig - \E\Big[2\xorig^\top\x\Big] - \left(\w^\top\xorig\right)^2
     + \\&\quad 2\left(\xorig^\top\w\right)\E\Big[\w^\top\x\Big] + \E\Big[\x^\top\x\Big] - \E\Big[\left(\x^\top\w\right)^2\Big]\\
    &= \xorig^\top\xorig - 2\xorig^\top\xorig - \left(\w^\top\xorig\right)^2
     + \\&\quad 2\left(\xorig^\top\w\right)^2 + \xorig^\top\xorig + \frac{\dimsym\epsilon^2}{3} - \left(\w^\top\xorig\right)^2 - \frac{\epsilon^2}{3} \\
     &= \frac{\epsilon^2(\dimsym - 1)}{3}
\end{split}
\end{equation}
which concludes the proof.
\end{IEEEproof}

\item
\begin{IEEEproof}[\refcorollary{corollary:individualfairness:uniformnoise:linearbinaryclassifier}]
Plugging the expectation from~\reftheorem{theorem:individualfairness:uniformnoise:linearbinaryclassifier} into Markov's inequality yields the claimed bound:
\begin{equation}
\begin{split}
    \prob\Big(\dist(\xcf, \xcfNew) \geq \delta\Big) &\leq \frac{{\E}\big[\dist(\xcf,\xcfNew)\big]}{\delta} \\
    &= \frac{\delta\epsilon^2(\dimsym - 1)}{3}
\end{split}
\end{equation}
\end{IEEEproof}

\end{enumerate}

\newpage
\section{Additional Plots}\label{sec:appendix:experiments:plots}
\begin{figure*}
  \caption{Wine data set: \emph{Median} $l_1$ distance between counterfactual of original sample and perturbed sample (using feature masking~\refeq{eq:perturbation:featuremasking}) for closest and plausible counterfactual explanations - for different number of masked features. Smaller values are better.}
  \label{fig:exp:perturbation:featuremasking:wine}
 \begin{minipage}[b]{0.3\textwidth}
    \includegraphics[width=\textwidth]{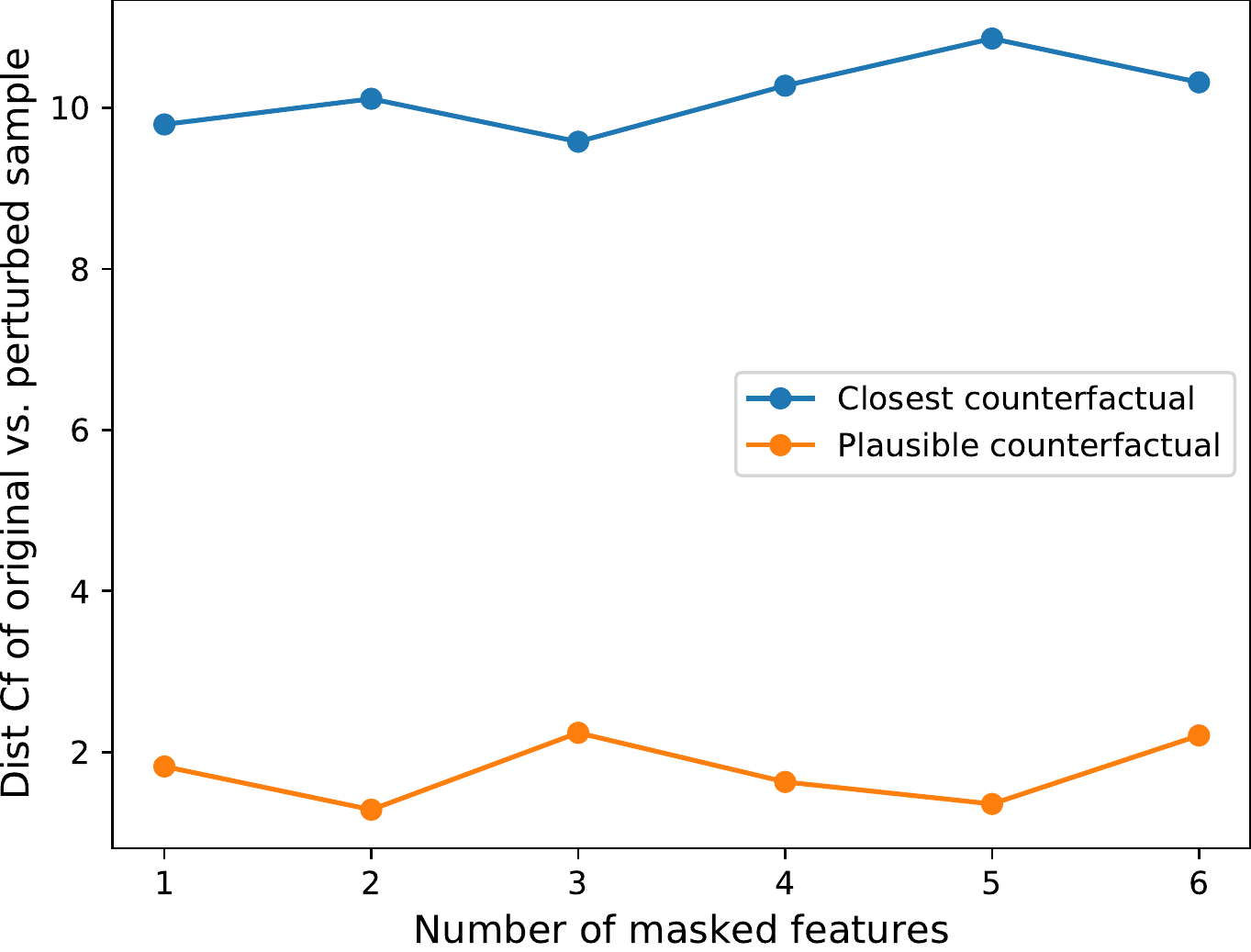}  
    \caption*{Softmax regression} 
   \end{minipage}
  \hfill
  \begin{minipage}[b]{0.3\textwidth}
    \includegraphics[width=\textwidth]{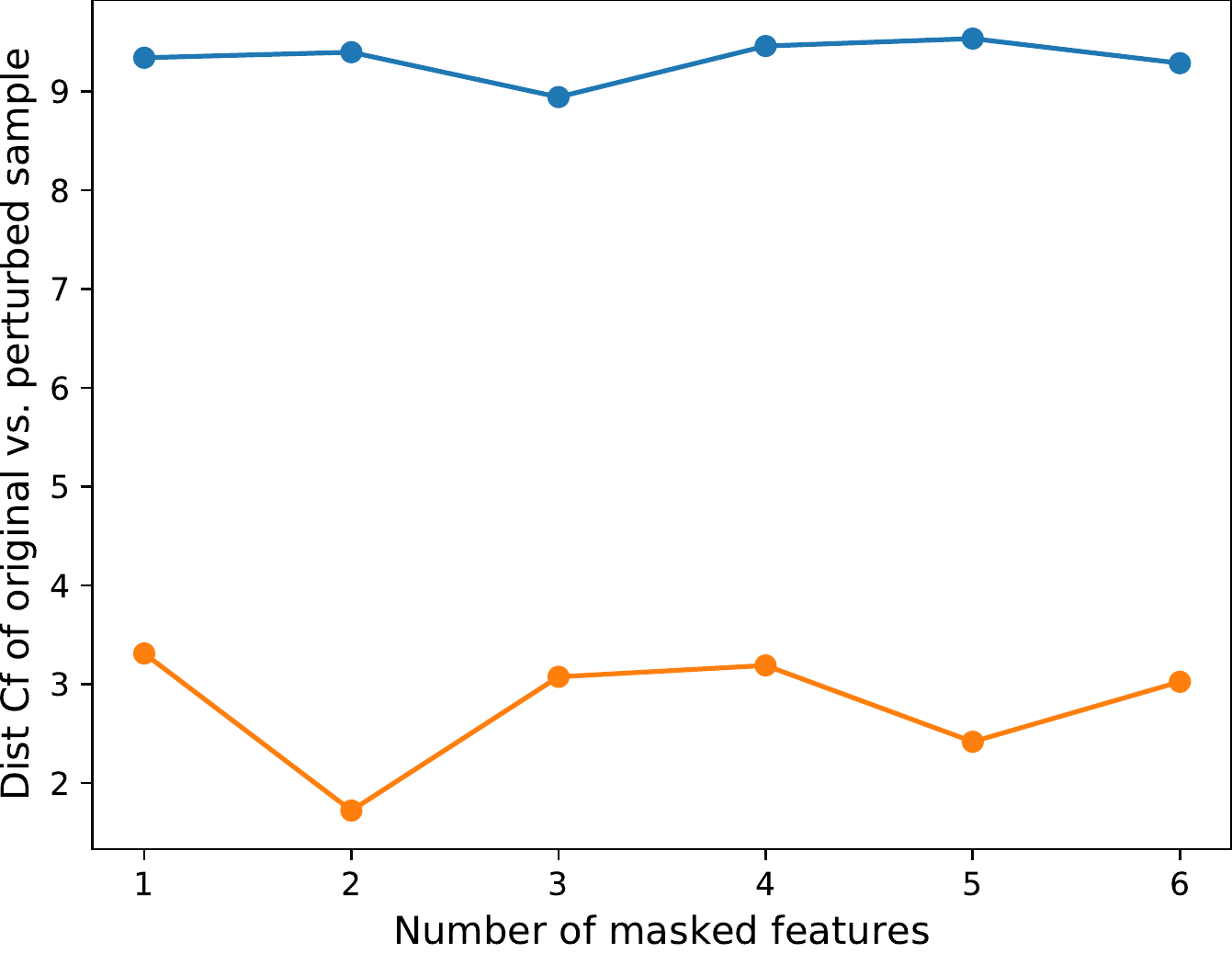}
    \caption*{Decision tree} 
  \end{minipage}
  \hfill
  \begin{minipage}[b]{0.3\textwidth}
    \includegraphics[width=\textwidth]{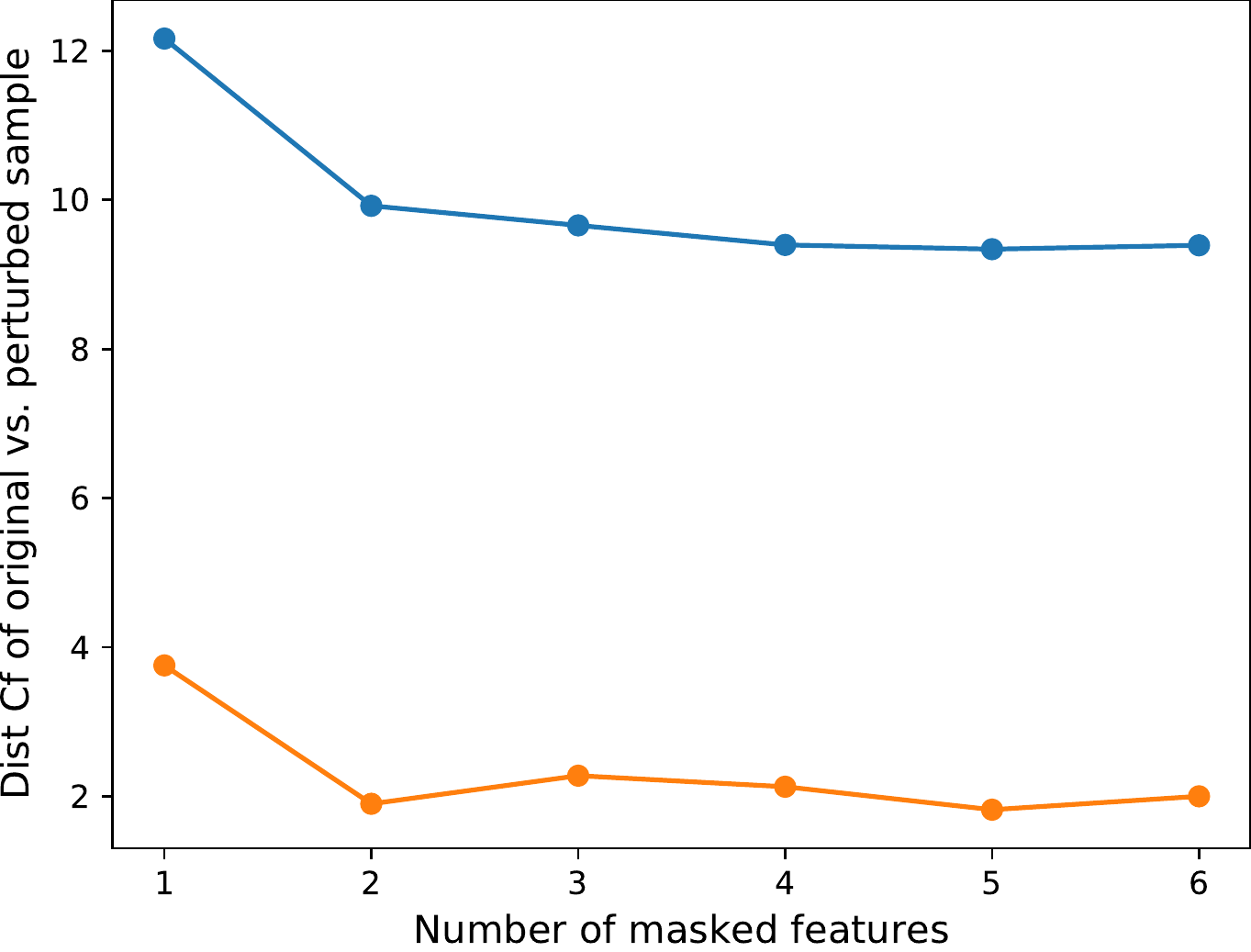}
    \caption*{GLVQ} 
 \end{minipage}
\end{figure*}
\begin{figure*}
  \caption{Breast cancer data set: \emph{Median} $l_1$ distance between counterfactual of original sample and perturbed sample (using feature masking~\refeq{eq:perturbation:featuremasking}) for closest and plausible counterfactual explanations - for different number of masked features. Smaller values are better.}
  \label{fig:exp:perturbation:featuremasking:breastcancer}
 \begin{minipage}[b]{0.3\textwidth}
    \includegraphics[width=\textwidth]{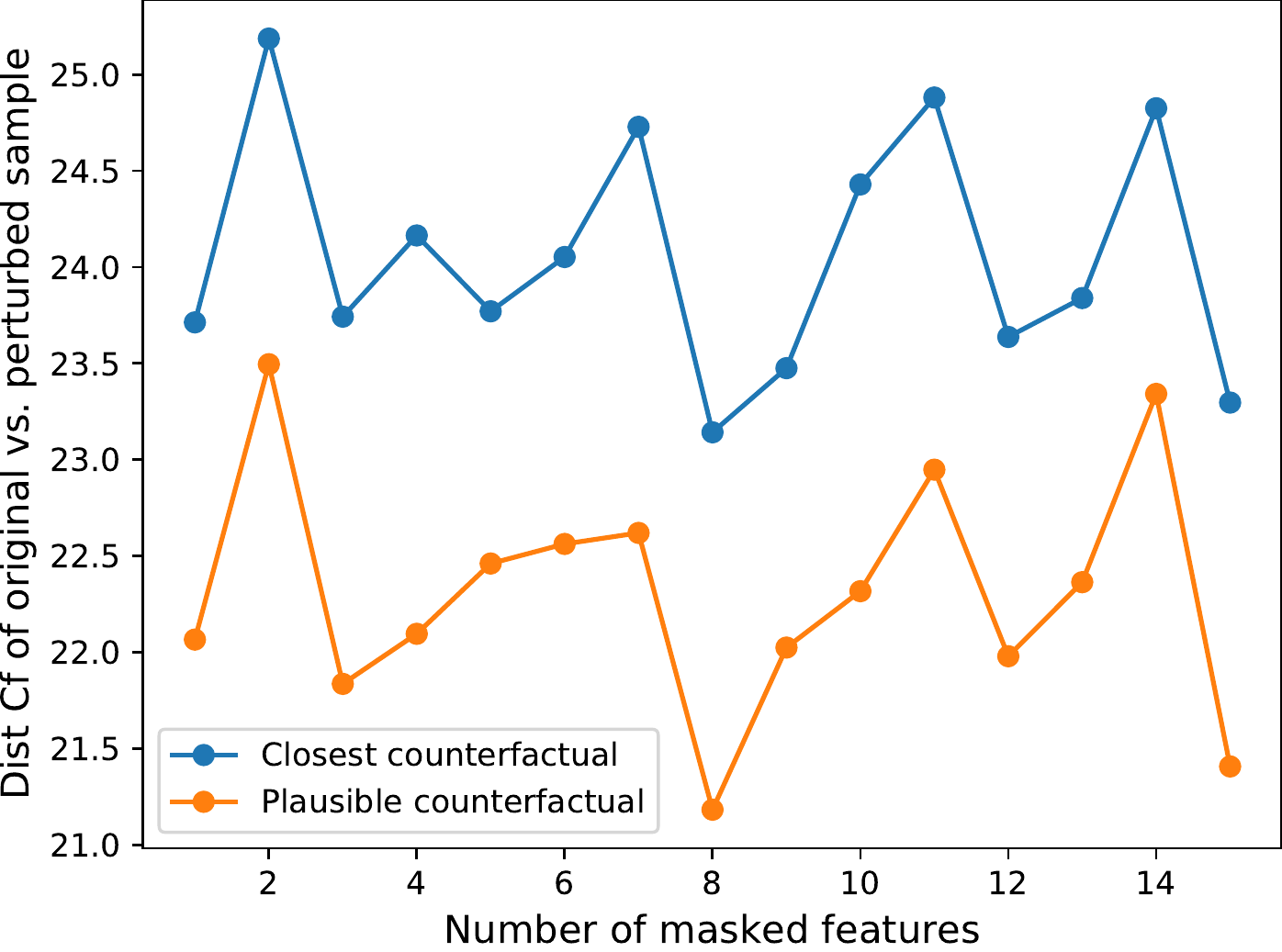}  
    \caption*{Softmax regression} 
   \end{minipage}
  \hfill
  \begin{minipage}[b]{0.3\textwidth}
    \includegraphics[width=\textwidth]{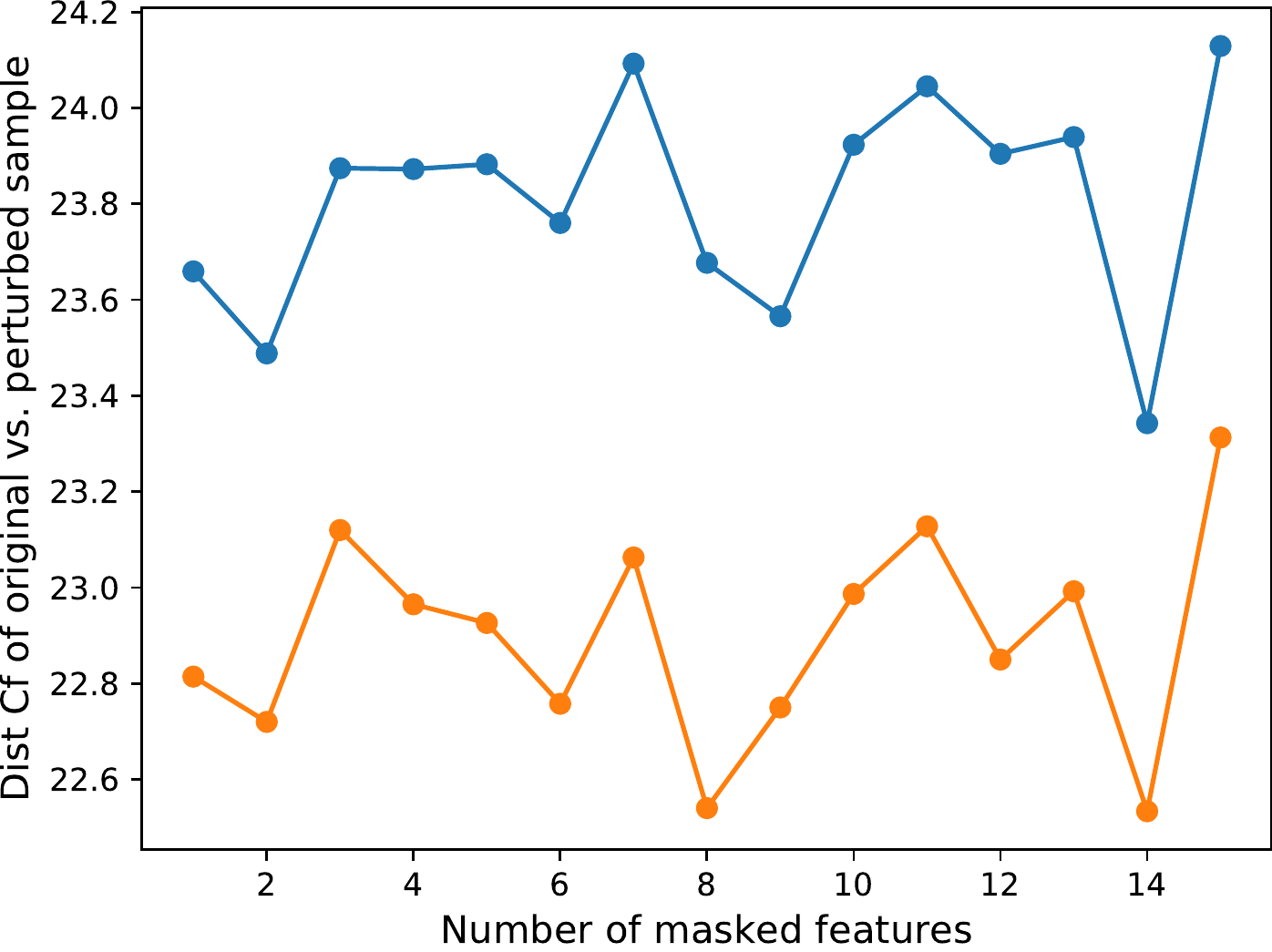}
    \caption*{Decision tree} 
  \end{minipage}
  \hfill
  \begin{minipage}[b]{0.3\textwidth}
    \includegraphics[width=\textwidth]{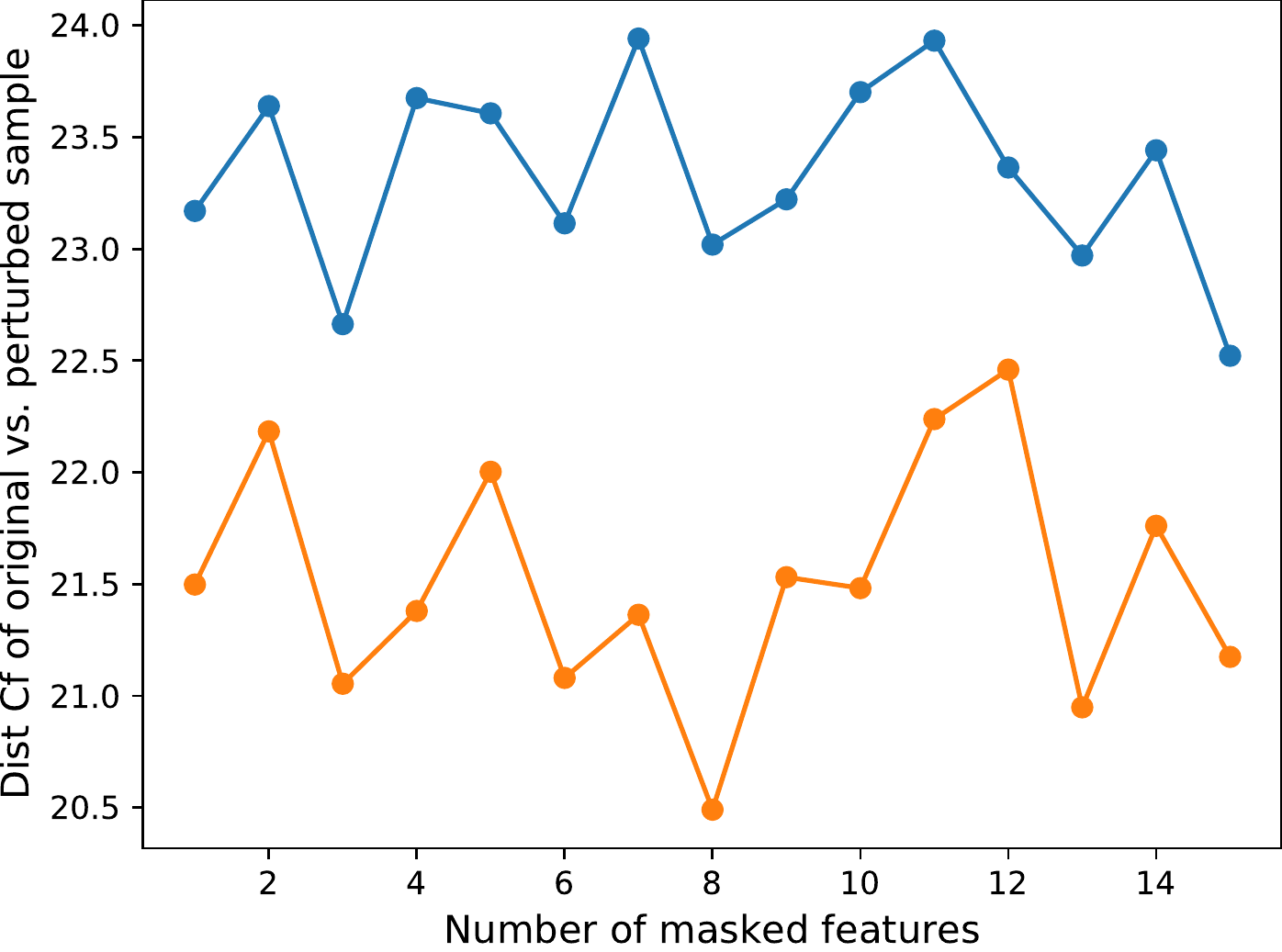}
    \caption*{GLVQ} 
 \end{minipage}
\end{figure*}

\end{document}